\documentclass[journal]{IEEEtran}

\usepackage{amsmath}
\usepackage{graphicx}
\usepackage{subfig} 
\usepackage{tabu}  

\usepackage[colorlinks,linkcolor=blue,anchorcolor=blue,citecolor=green]{hyperref}
\usepackage{multirow}
\usepackage{dsfont}
\usepackage{stfloats}
\usepackage{amsmath}
\usepackage{cite}
\usepackage{booktabs}
\usepackage{algorithm}
\usepackage{algorithmic}
\usepackage{amssymb}
\usepackage{makecell}
\usepackage{caption}
\ifCLASSINFOpdf

\else
 
\fi

\begin{document}

\captionsetup{font = {small}}

\title{Unsupervised Multimodal Change Detection Based on Structural Relationship Graph Representation Learning}

\author{
        Hongruixuan~Chen,~\IEEEmembership{Student Member,~IEEE,}
        Naoto~Yokoya,~\IEEEmembership{Member,~IEEE,}
        Chen~Wu,~\IEEEmembership{Member,~IEEE,}
        and~Bo~Du,~\IEEEmembership{Senior~Member,~IEEE}

\thanks{Manuscript submitted October 1, 2022.}
\thanks{H. Chen is with Graduate School of Frontier Sciences, the University of Tokyo, Chiba, Japan (e-mail: Qschrx@gmail.com).}
\thanks{N. Yokoya is with Graduate School of Frontier Sciences, the University of Tokyo, 277-8561 Chiba, Japan, and also with the Geoinformatics Unit, RIKEN Center for Advanced Intelligence Project (AIP), RIKEN, 103-0027 Tokyo, Japan. (e-mail: yokoya@k.u-tokyo.ac.jp).}
\thanks{C. Wu is with the State Key Laboratory of Information Engineering in Surveying, Mapping and Remote Sensing, Wuhan University, Wuhan, P.R. China (e-mail: chen.wu@whu.edu.cn).}
\thanks{B. Du is with the School of Computer Science, and Collaborative Innovation Center of Geospatial Technology, Wuhan University, Wuhan, P.R. China (email: dubo@whu.edu.cn).}
}

\markboth{SUBMITTED TO IEEE TRANSACTIONS ON GEOSCIENCE AND REMOTE SENSING ON October 1, 2022}%
{Shell \MakeLowercase{\textit{\emph{et al.}}}: Bare Demo of IEEEtran.cls for IEEE Journals}

\maketitle

\begin{abstract}
  Unsupervised multimodal change detection is a practical and challenging topic that can play an important role in time-sensitive emergency applications. To address the challenge that multimodal remote sensing images cannot be directly compared due to their modal heterogeneity, we take advantage of two types of modality-independent structural relationships in multimodal images. In particular, we present a structural relationship graph representation learning framework for measuring the similarity of the two structural relationships. Firstly, structural graphs are generated from preprocessed multimodal image pairs by means of an object-based image analysis approach. Then, a structural relationship graph convolutional autoencoder (SR-GCAE) is proposed to learn robust and representative features from graphs. Two loss functions aiming at reconstructing vertex information and edge information are presented to make the learned representations applicable for structural relationship similarity measurement. Subsequently, the similarity levels of two structural relationships are calculated from learned graph representations and two difference images are generated based on the similarity levels. After obtaining the difference images, an adaptive fusion strategy is presented to fuse the two difference images. Finally, a morphological filtering-based postprocessing approach is employed to refine the detection results. Experimental results on five datasets with different modal combinations demonstrate the effectiveness of the proposed method.
\end{abstract}


\IEEEpeerreviewmaketitle

\section{Introduction}\label{sec:1}

\IEEEPARstart{C}{hange} detection is the process of identifying the changes of objects or phenomena in the same geographical area by analyzing remote sensing images acquired at different times \cite{Singh1989}. It has been widely used in many real-world applications, such as urban studies, ecosystem monitoring, resource management, armed conflict monitoring, and damage assessment \cite{Desclee2006,Brunner2010,Wu2017b,Luo2018,Wu2021a,Chen2022Dual}. 

\par Unimodal change detection or homogeneous change detection, where pre-change and post-change images are collected by the same kind of sensors, has been widely studied over the past decades. Many mature and effective paradigms, including traditional and deep learning-based models, have been proposed for different kinds of remote sensing images, including multispectral images \cite{Sharma2007,Wu2014,Du2019a,Mou2019}, hyperspectral images \cite{Liu2019,Liu2015a,Liu2015b,Wang2019,Hu2021Hyperspectral}, very-high-resolution optical images \cite{Chen2019Deep,Saha2019,Peng2020,Wu2021unsupervised,guo2021deep,Tang2022}, and synthetic aperture radar (SAR) images \cite{Gong2012,Gong2016,Gao2016,Yang2019a,Sun2020}. Recently, with the rapid development of earth observation technology, more and more remote sensing images representing land-cover information can be obtained from different sensors at the same time. This development provides data support for the research of multimodal change detection. Compared to unimodal change detection, the pre-change and post-change images in multimodal change detection are obtained by different sensors. This means that multimodal change detection can alleviate the constraints of atmospheric conditions and revisit period cycles of satellites, thereby providing land-cover change information timely. Thus, multimodal change detection has great practical significance, especially for immediate evaluation and emergency disasters.  

\par However, little research has been conducted on multimodal change detection despite its significance. Since the pre-change and post-change images acquired by different sensors have different modalities, like optical and SAR images, it is difficult to get accurate detection results from multimodal images by those intuitive paradigms designed for unimodal images. Generally, multimodal change detection can be divided into supervised and unsupervised ones depending on whether or not label information is provided for the detection model. However, annotating labels for multimodal data is very labor-intensive and requires extensive expert knowledge as a guide \cite{Chen2019a,Wu2021multiscale}. Thus, unsupervised methods are more popular in practice, and undoubtedly more challenging. In this paper, we focus on unsupervised multimodal change detection.

\par The fundamental idea of unsupervised multimodal change detection is to find a domain, where incomparable multimodal images become comparable. According to the transformation way and domain type, the existing unsupervised method can be generally divided into four categories: Classification methods, modality translation methods, feature learning-based methods, and similarity measure-based methods.

\par 1) Classification methods transform the multimodal images into a common category domain by classifying the pre-change and post-change images separately, and then compare the classification maps in the category domain to detect the changes. Wan \emph{et al.} \cite{Wan2019Post} presented a post-classification comparison method for SAR and optical image change detection. They first generated objects for multimodal images and then compound classification was carried out at the object level. Based on this work, they further proposed a region-based multitemporal hierarchical Markov random field to improve classification accuracy \cite{Wan2019An}. As can be seen, this type of method relies heavily on classification accuracy. However, unsupervised classification methods are difficult to obtain accurate classification results.

\par 2) Modality translation methods reduce the modality difference by transforming one image from its modality to the modality of the other image. The paradigms of style transfer in the field of computer vision are introduced and developed. In \cite{Niu2019}, a conditional generative adversarial network (GAN) was applied to translate the optical image to the style of the SAR image. An approximate network was further designed to reduce the pixel-wise difference between the SAR image and translated images. Jiang \emph{et al.} \cite{Jiang2020Change} proposed a deep homogeneous feature fusion model based on image style transfer. The optical images were transferred to the style of the SAR images, and change detection was performed. In \cite{Luppino2022Code} and \cite{Luppino2022Deep}, the idea of adversarial learning, image-to-image translation, and cycle consistency were introduced for designing networks and transforming image modality.  Nevertheless, the running time and computational overhead of style transfer learning approaches and GANs are quite large.

\par 3) Feature learning-based methods design an appropriate model, mainly deep learning models, to find a high-dimensional feature space where features of multimodal images can be directly compared. Zhang \emph{et al.} \cite{Zhang2016Change} proposed to utilize the denoising autoencoder to learn high-level representations from image patches for multiresolution images. In \cite{Liu2018}, an unsupervised symmetric convolutional coupling network (SCCN) was proposed for multimodal change detection. In SCCN, multimodal images could be transformed into a feature space where their feature representations become more consistent. In \cite{Zhao2017a}, the deep belief network (DBN) was introduced to learn features. Zhan \emph{et al.} \cite{Zhan2018b} proposed an iterative feature mapping network to learn multi-class change types from multimodal remote sensing images. In \cite{Zhan2018Log}, Zhan \emph{et al.} further used the logarithmic transformation to transform SAR images so that they have similar statistical distributions as the optical images. Recently, Wu \emph{et al.} \cite{Wu2021b} proposed a commonality autoencoder for commonalities exploration, which can discover common features by transforming heterogeneous image representations. Due to the great learning ability of deep learning models, these methods have achieved good performance on the multimodal change detection task. These methods assume that unchanged areas occupy a large proportion of multimodal image pairs. Under this circumstance, they can directly shrink the distance between paired features to learn the common feature space. However, if the changed areas occupy a large proportion, the learning process will be affected, thereby degrading the accuracy of the results.

\par 4) Similarity measure-based methods define a modality-independent metric that can be used to distinguish the changed and unchanged areas. In \cite{Wan2018Multi}, the distance of the sorted histogram was generated within the image and the dissimilarity between the multimodal images was estimated by this measure. Liu \emph{et al.} \cite{Liu2018a} presented a homogeneous pixel transformation method to detect changes between panchromatic and multispectral images. For each pixel in the pre-change/post-change image, they estimated its mapping pixel in the post-change/pre-change image using the K-nearest neighbor (KNN) algorithm. Luppino \emph{et al.} \cite{Luppino2019Unsupervised} proposed an affinity matrix distance to calculate the change possibility of each pixel, which can be directly used to generate the change map. Also, subsequent methods, like unsupervised image regression \cite{Luppino2019Unsupervised} and deep image translation-based methods \cite{Luppino2022Deep,Luppino2022Code}, can be employed to get more accurate detection results. In \cite{Mignotte2020}, the self-similarity property of images was introduced for multimodal change detection. The pre-change image was transformed into the domain of the post-change image by fractal projection according to self-similarity. Then, the difference image was obtained by comparing the transformed image with the post-change image. Also exploring self-similarity, Sun \emph{et al.} \cite{Sun2021bPatch} presented a patch similarity graph matrix (PSGM) and multiplied multimodal images with the PSGM for image regression. They also applied self-similarity to construct graphs, called NPSG \cite{Sun2021c}, representing the structures for each image, and built the similarity relationships between heterogeneous images. In \cite{Sun2021a}, they designed an iterative robust graph framework to reduce the computational overhead of NPSG and achieve better detection performance.

\par In this paper, we follow the research line of the similarity measure-based method. The main advantages of these methods are that they are intuitive and easy to implement in practice. However, the similarity measures proposed in these methods only utilize low-level information in remote sensing images, which are not robust across different modal combinations of pre-change and post-change images. Consequently, these methods often face performance degradation when detection conditions are complex, like the large diversity of land-cover objects, and the strong speckle noise in SAR images.  

\par Therefore, we propose an unsupervised structural relationship graph learning framework for multimodal change detection. The proposed framework not only employs self-similarity (called nonlocal structural relationship in this paper) but also explores the local structural relationship observed from multimodal remote sensing images. Since the structural information of images can be expressed in the form of graph data, we naturally introduce graph representation learning and design a structural relationship graph convolutional autoencoder (SR-GCAE). The proposed network can learn representative and robust graph features from information contained in vertices and edges of graphs through two reconstruction optimization objectives. Benefiting from this, the degree of local and nonlocal structural similarity can be calculated from the features learned by SR-GCAE instead of calculating it from low-level spectral features. Additionally, a simple but effective adaptive fusion strategy is presented to fuse the difference images obtained by calculating the similarity levels of two structural relationships.

\par In particular, the main contribution of our work can be summarized as: 
\begin{enumerate}
    \item A local structural relationship and a nonlocal structural relationship are simultaneously explored for unsupervised multimodal change detection. 
    \item We present the first attempt at designing a graph representation learning framework for unsupervised multimodal change detection. With two kinds of reconstruction objectives as the loss function, the proposed network can learn robust high-level graph representations to measure the similarity levels of local and nonlocal structural relationships.
    \item An adaptive fusion strategy based on the discrimination of change intensity in difference images is proposed to better highlight changed pixels and suppress unchanged pixels. 
    \item In five change detection datasets with different modality combinations, the proposed method outperforms the state-of-the-art methods, showing its superiority.      
\end{enumerate}

\par The remainder of this paper is organized as follows. In Section \ref{sec:2}, the preliminaries of two structural relationships and graph convolutional networks are introduced. Section \ref{sec:3} elaborates on the multimodal change detection framework based on the proposed SR-GCAE. To evaluate the proposed method, the experiments on five multimodal datasets are carried out in Section \ref{sec:4}. Finally, Section \ref{sec:5} draws the conclusion of our work in this article.

\section{Preliminaries}\label{sec:2}
\subsection{Structural Relationship in Multimodal Data}
\par Due to the large difference in modality, it is difficult to accurately detect changes by comparing multimodal remote sensing images directly in the original spectral domain. Therefore, unsupervised multimodal change detection is essentially about finding a domain that makes multitemporal images with modal heterogeneity comparable. In this work, we  exploit the local and nonlocal structural relationships in multimodal remote sensing images. 

\par
\begin{figure}[!t]
  \centering
  \subfloat[]{
    \includegraphics[width=1.6in]{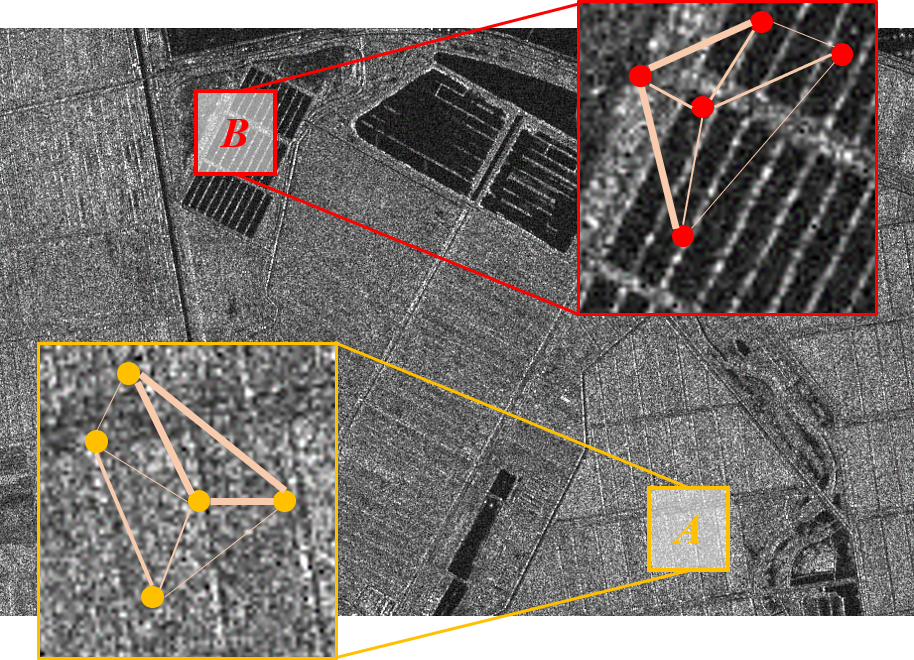}
  \label{fig_first_case}}
  \hfil
  \subfloat[]{
    \includegraphics[width=1.6in]{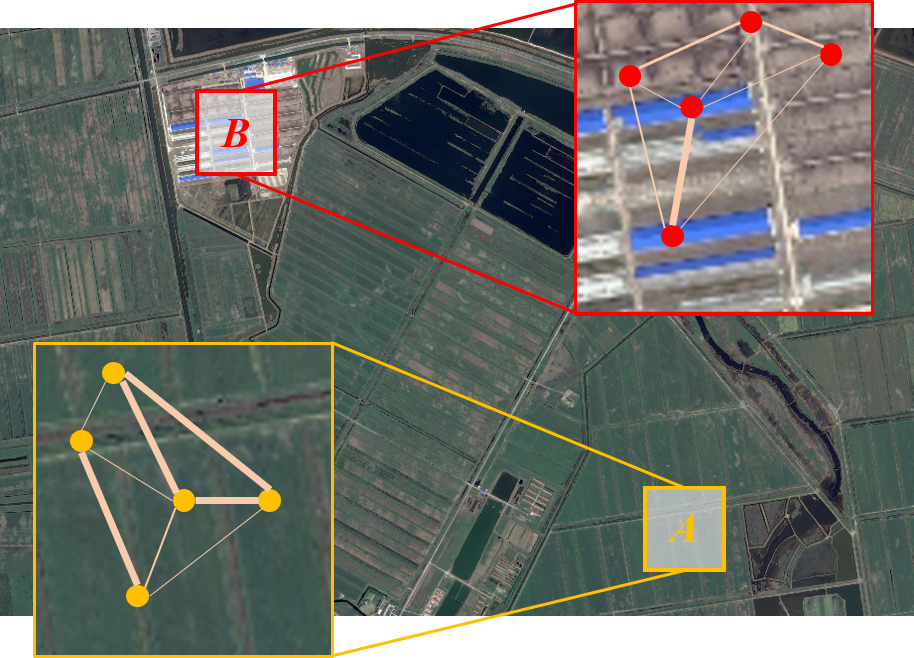}
  \label{fig_second_case}}
  \caption{Local structural relationship in multimodal remote sensing images. (a) Pre-change image. (b) Post-change image. Here, the point means pixel, and the line represents the relationship between pixels, the thicker the line, the more similar the pixels are to each other.}
  \label{Fig1_LocalSim}
\end{figure}

 Fig. \ref{Fig1_LocalSim} shows a pair of multimodal remote sensing images and two local areas \textbf{\emph{A}} and \textbf{\emph{B}} within them. The pre-change image is a SAR image and the post-change image is an optical image. Although pre-change and post-change images show a large modality difference, the relationship between inner pixels of area \textbf{\emph{A}} should be similar in both pre-change and post-change images, as its corresponding land-cover objects are the same. On the other hand, the relationship between a large number of inner pixels of area \textbf{\emph{B}} cannot retain consistency between pre-change and post-change images as a result of the occurring changes in land-cover objects. We refer to this relationship as the local structural relationship.

\par
\begin{figure}[!t]
  \centering
  \subfloat[]{
    \includegraphics[width=1.6in]{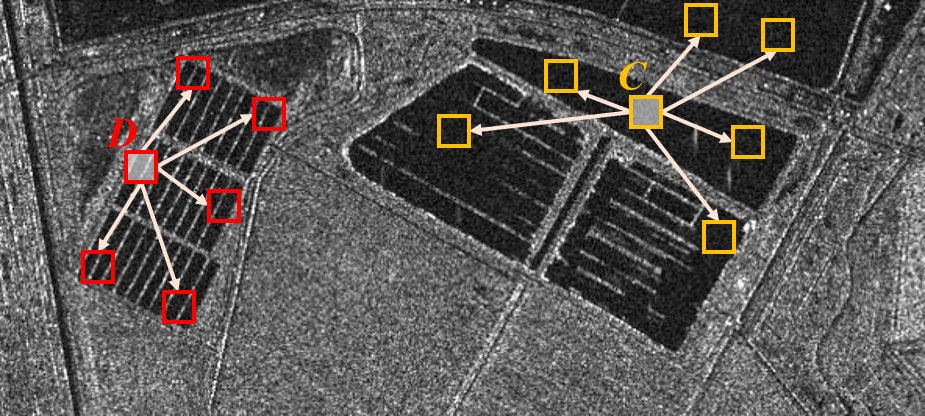}
  \label{fig_first_case}}
  \hfil
  \subfloat[]{
    \includegraphics[width=1.6in]{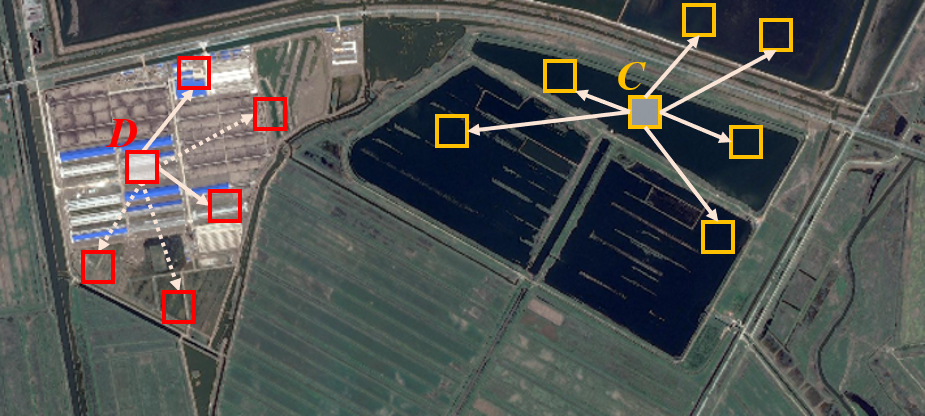}
  \label{fig_second_case}}
  \caption{Nonlocal structural relationship in multimodal remote sensing images. (a) Pre-change image. (b) Post-change image. Here, the rectangular boxes represent small areas of the image and the lines represent the relationship between image areas, with solid lines indicating that the image areas are similar to each other and dashed lines indicating that they are not.}
  \label{Fig2_NLocalSim}
\end{figure}

 The second relationship is based on the self-similarity property of images, which has been widely used in image denoising \cite{Buades2005non,dabov2007image,buades2010image}. According to the self-similarity property, for each small area in the image, some similar areas can be founded within the same image. In Fig. \ref{Fig2_NLocalSim}, for two areas \textbf{\emph{C}} and \textbf{\emph{D}} in the pre-change image, we can find some similar areas for them. Since area \textbf{\emph{C}} is unchanged, the relationship between area \textbf{\emph{C}} and its similar areas can preserve nearly consistent in the post-change image. However, for area \textbf{\emph{D}}, this relationship cannot be retained because the change event has occurred. We call this relationship as the nonlocal structural relationship.

\par Since these two relationships are measured within the image, they can be seen as modality-independent. This means that we can use the similarity degree of these two relationships as an indirect measure of the changes. The next questions, then, are how to represent these two relationships in an appropriate model and how to use them to detect changes.

\subsection{Graph Convolutional Networks}
\par In our method, the graph model is employed to represent structural information contained in multimodal remote sensing images. Then, we present a structural relationship graph convolutional autoencoder to learn the information represented by graphs for multimodal change detection. Thus, we briefly introduce basic concepts of graph and graph convolutional networks as the preliminaries of our method.

\par
\begin{figure}[t]
  \centering
  \includegraphics[width=3.1in]{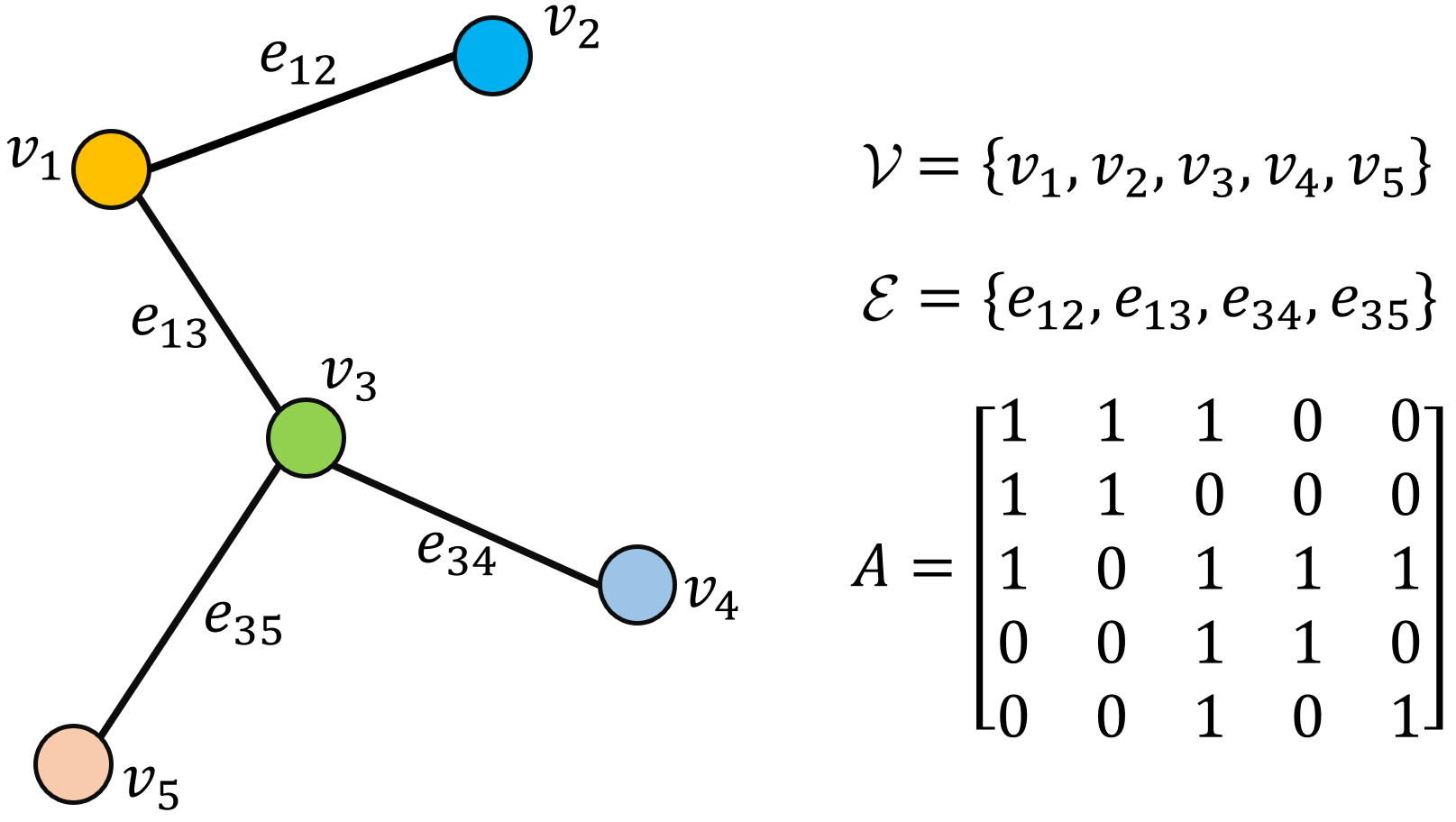}
  \caption{An illustration of a graph.}
  \label{Fig3_Graph}
\end{figure}

 As shown in Fig. \ref{Fig3_Graph}, a graph is a non-Euclidean data structure consisting of vertices (or nodes) and edges, which can be represented as $\mathcal{G}=\left(\mathcal{V}, \mathcal{E}\right)$ where $\mathcal{V}$ is the set of vertices with the number $\left| \mathcal{V} \right|=N$ and $\mathcal{E}$ is the set of edges. Let $v_i\in \mathcal{V}$ denote a vertex and $e_{ij}=\left(v_i,v_j\right)\in \mathcal{E}$ denote an edge. Then, adjacency matrix $A$ can be defined as an $N\times N$ matrix with $A_{ij}=1$ if $e_{ij}\in \mathcal{E}$ and $A_{ij}=0$ if $e_{ij}\ \notin \mathcal{E}$. Based on $A$, the graph Laplacian matrix $L$ can be defined:
\begin{equation}
  L=D-A,
\label{eq:1}
\end{equation}
where $D$ is a diagonal matrix of $A$, namely $D_{ii}=\sum_{j=1}^{N}A_{ij}$. 
\par We can further calculate the normalized Laplacian matrix $\tilde{L}$ for enhancing the generalization ability of the graph:
\begin{equation}
 \tilde{L}=I-D^{-\frac{1}{2}}AD^{-\frac{1}{2}},
\label{eq:2}
\end{equation}
where $I$ is the identity matrix.
\par As an emerging network architecture, graph convolutional networks can effectively handle graph structure data by modeling relations between vertices. Now, there are many variants of graph convolutional networks \cite{kipf2016semi,defferrard2016convolutional,velivckovic2017graph,wu2019simplifying,park2019symmetric}. Our work follows the standard graph convolutional network proposed in \cite{kipf2016semi}, which is essentially a first-order approximation of localized spectral convolution on graphs.

\par Given a signal $s\in \mathbb{R}^{N}$ (a scalar for each vertex) and a filter $g_\theta=\mathrm{diag}\left(\theta\right)$ parameterized by $\theta\in \mathbb{R}^N$, the spectral convolution of $s$ and $g_\theta$ can be performed by decomposing $s$ on the Fourier domain and then multiplying each frequency by $g_\theta$ as
\begin{equation}
 g_\theta\star s=Ug_\theta U^\top s,
\label{eq:3}
\end{equation}
where $U$ is the matrix of eigenvectors of the normalized Laplacian matrix $\tilde{L}=I-D^{-\frac{1}{2}}AD^{-\frac{1}{2}}=U\mathrm{\Lambda}U^\top$. $\mathrm{\Lambda}$ is the diagonal matrix of eigenvalues of $\tilde{L}$. $U^\top s$ denotes the graph Fourier transform of $s$. $g_\theta$ can be understood as a function of the eigenvalues of $\tilde{L}$, i.e., $g_\theta\left(\mathrm{\Lambda} \right)$. However, evaluating Eq. (\ref{eq:3}) requires explicitly calculating the Laplacian eigenvectors, which is not computationally feasible for large-scale graph data. To solve this problem, a feasible way is to approximate the filter $g_\theta$ by the Chebyshev polynomials up to the $K$-th order \cite{hammond2011wavelets}:
\begin{equation}
 g_{\theta}\star s\approx\sum_{k=0}^{K}{\theta_k^\prime T_k\left(\tilde{L}\right)s },
\label{eq:4}
\end{equation}
where $T_{k}$ is the Chebyshev polynomials.
\par Going back to the graph convolutional network involved in our work, it limits $K=1$ and approximates the largest eigenvalue of $\tilde{L}$ as $\lambda_{max}\approx2$. By doing so, Eq. (\ref{eq:4}) can be further simplified to:
\begin{equation}
 g_\theta\star s\approx\theta\left(I+D^{-\frac{1}{2}}AD^{-\frac{1}{2}}\right)s.
\label{eq:5}
\end{equation}
\par Based on Eq. (\ref{eq:5}), there is the following propagation rule of the graph convolutional layer:
\begin{equation}
 H^{\left(l+1\right)}=\sigma\left(\tilde{D}^{-\frac{1}{2}} \tilde{A} \tilde{D}^{-\frac{1}{2}}H^{\left(l\right)}W^{\left(l\right)}\right).
\label{eq:6}
\end{equation}
Here, $\ H^{\left(l\right)}$ and $H^{\left(l+1\right)}$ are the input and output of the $l$-th graph convolutional layer, $W^{\left(l\right)}$ denotes the layer trainable weights, and $\sigma(.)$ is the activation function to introduce nonlinearity for the learned features. 

\section{Methodology}\label{sec:3}
\par
\begin{figure*}[ht]
  \centering
  \includegraphics[width=7.0in]{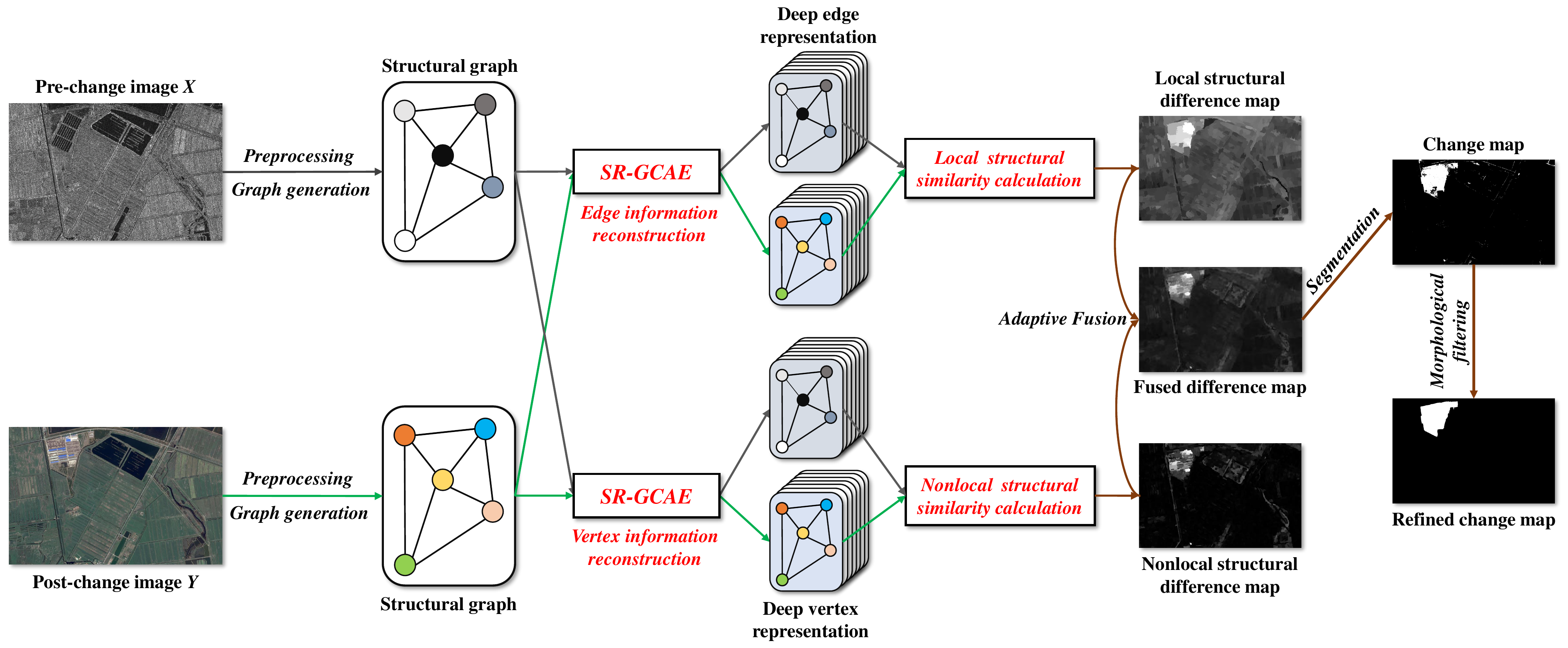}
  \caption{Overview of unsupervised multimodal change detection based on structural relationship graph representation learning.}
  \label{fig:SRGCAE_CD}
\end{figure*}

 In this section, we present an unsupervised multimodal change detection framework based on the aforementioned two structural relationships and graph convolutional networks, as shown in Fig. \ref{fig:SRGCAE_CD}. Firstly, multimodal images are preprocessed by geometric alignment and image normalization. Then, we construct graphs to represent the crucial structural information in multimodal images. After that, we present the structural relationship graph convolutional autoencoder and employ it on the constructed graphs to learn structural information. Our framework has two SR-GCAEs. The first SR-GCAE learns the edge information of graphs, which is then used to reflect the similarity degree of the local structural relationship between two images. The second network learns the vertex information of graphs, which is then used for nonlocal structural similarity measurement. Using the deep graph representations learned by two SR-GCAEs, we can measure the change level and generate corresponding difference images. After that, a fusion strategy is proposed to adaptively fuse the two difference images. Finally, a simple postprocessing step based on morphological filtering is used to refine the change map.

\subsection{Data Preprocessing}
\par Given a pair of multimodal remote sensing images acquired at times $T_1$ and $T_2$, we denote the pre-change image with modality $\mathcal{X}$ as $X\in \mathbb{R}^{H_{X}\times W_{X}\times C_{X}}$ and the post-change image with modality $\mathcal{Y}$ as $Y\in \mathbb{R}^{H_{Y}\times W_{Y}\times C_{Y}}$, where $H_{X}$,$W_{X}$,$C_{X}$ and $H_{Y}$,$W_{Y}$,$C_{Y}$ are the height, width, and number of channels of pre-change image and post-change image, respectively. The pixels in both images are denoted as $x(h,w,c)$ and $y(h,w,c)$, respectively. 

\par The first step of preprocessing is image alignment. It is the process of aligning two or more remote sensing images of the same scene acquired at different times \cite{Zitova2003}. Given a pair of multimodal remote sensing images, change detection between the corresponding regions is only meaningful if the two images are geometrically aligned. In our framework, there are four main steps in image alignment: collecting matched point pairs, building a transformation model, transforming the image and resampling. As the spatial resolutions of multimodal remote sensing images are generally different, the resampling step is to resample images with a relatively higher resolution to the reference image with a relatively lower resolution. With the four steps described above, a given multimodal image pair is geometrically aligned. We denote the co-aligned images as $\hat{X}\in R^{H\times W\times C_X}$ and $\hat{Y}\in R^{H\times W\times C_Y}$. 

\par The second step of preprocessing is image normalization. For unimodal change detection, the major purpose of normalization is to eliminate the radiometric difference between multitemporal images caused by different imaging conditions \cite{Zhang2014}. For multimodal change detection, image normalization is also meaningful because it can assign the value of various dimensions of the input multimodal images to be in a similar range. It is beneficial for the performance of the subsequent steps. Optical and SAR images are the two most common types of images used for change detection tasks. The multimodal images used in the experiments of this work belong to these two types. Therefore, we will only present here the normalization methods for these two types of images.
For optical images, we directly normalize their pixel values to the range $[0, 1]$:
\begin{equation}
 \tilde{x}\left(h,w,c\right)=\frac{\hat{x}\left(h,w,c\right)-min_c}{max_c-min_c},
\label{eq:7}
\end{equation}
where $max_c$ and $min_c$ are the maximum and minimum pixel values of the image in the $c$-th band. 

\par For SAR images, we first take logarithmic operations to suppress the speckle noise and then normalize the logarithmic transformed image:
\begin{equation}
\left\{
    \begin{aligned}
      &\hat{x}^{log}(h,w,c)=\log(1+\hat{x}(h,w,c)),  \\
      &\tilde{x}(h,w,c)=\frac{\hat{x}^{log}(h,w,c)-min_c}{max_c-min_c}.
    \end{aligned}
  \right.
\label{eq:8}
\end{equation}

\par The normalized two images are denoted as $\tilde{X}\in R^{H\times W\times C_X}$ and $\tilde{Y}\in R^{H\times W\times C_X}$. 

\subsection{Structural Graph Construction}

\par The second step is to construct structural graph data to represent the structural information of multimodal remote sensing images. To construct the graph data, the basic analysis unit in our method is not a pixel but a small part of the image. Image patches are an optional and common way to represent the structural information of an image \cite{Gong2016,Li2019a,Sun2021c}. However, generating image patches by sliding windows is a time-consuming process, and the distribution of land-cover objects is often not square. With these in mind, we apply image segmentation algorithms to obtain image objects (also called superpixels) as the basic units. They can better reflect the image structures than square image patches and are generated in a more efficient way.

\par In particular, the fractal net evolution approach (FNEA) \cite{baatz2000multi} is employed to get the image objects. Compared to some segmentation methods proposed for RGB images such as SLIC \cite{Achanta2012SLIC}, FNEA has no band number restrictions on the input data and is more suitable for remote sensing images. The basic idea of the FNEA algorithm is to define and measure the heterogeneity between neighboring image objects, and to use this as a basis for whether to merge objects. The merging criterion is defined as follows: 
\begin{equation}
f=w_{channel}h_{channel}+w_{spatial}h_{spatial}<T,
\label{eq:9}
\end{equation}
where $f$ is the merging criteria based on the channel heterogeneity $h_{channel}$ and spatial heterogeneity $h_{spatial}$ as defined in \cite{baatz2000multi}; $w_{channel}$ and $w_{spatial}$ are the weight of corresponding heterogeneity; $T$ is a preset merging threshold. 

\par Due to the different modality and land-cover objects distribution, executing FNEA on $\tilde{X}$ and $\tilde{Y}$ separately leads to different segmentation results. In this paper, we adopt the co-segmentation strategy. The multimodal image pair is stacked in the channel dimension as a single image to get a unified segmentation result by FNEA. We denote the co-segmentation map as
\begin{equation}
\left\{
    \begin{aligned}
      &\mathrm{\Omega}=\{\mathrm{\Omega}\left(i\right) \mid i=1,2,\ldots,N_{cs}\};  \\
      &\mathrm{\Omega}\left(i\right)\cap\mathrm{\Omega}\left(j\right) =\varnothing \ \ \mathrm{if} \ i\ \neq j;  \\
      &\bigcup_{i=1}^{N_{cs}}\mathrm{\Omega}\left(i\right)=\{\left(h,w\right)\mid h=1,\dots,H; w=1,\ldots,W\}. 
    \end{aligned}
\right.
\label{eq:10}
\end{equation}
where $N_{cs}$ is the number of objects.

\begin{figure*}[!ht]
  \centering
  \includegraphics[width=7.0in]{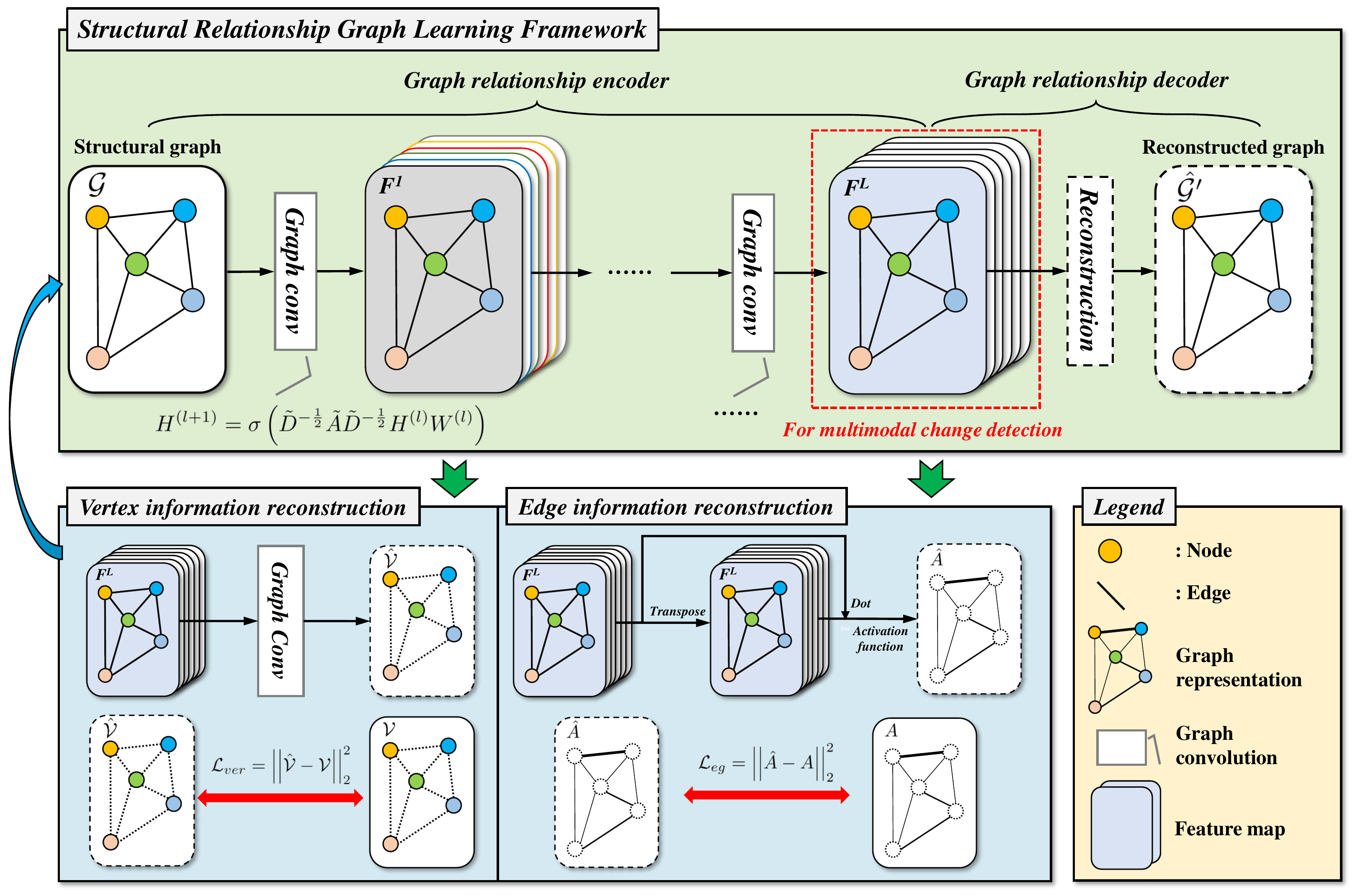}
  \caption{Illustration of structural relationship graph representation learning framework. Here, two optimization objectives are presented that make the SR-GCAE reconstruct the vertex information and edge information.}
  \label{fig:SRGCAE}
\end{figure*}

\par The $i$-th object of $\tilde{X}$ and $\tilde{Y}$ can be expressed as $O_i^{\tilde{X}}=\left\{\tilde{x}\left(h,w,c\right)\mid h,w\in\mathrm{\Omega}(i),c=1,2,...,C_X\right\}$ and $O_i^{\tilde{Y}}=\left\{\tilde{y}\left(h,w,c\right) \mid h,w\in\mathrm{\Omega}(i),c=1,2,...,C_Y\right\}$, respectively. Based on the generated objects, we can construct the graph data to represent the structural information. For pre-change image $X$, the structural information contained in $O_i^{\tilde{X}}$ can be represented by a graph as
\begin{equation}
\left\{
    \begin{aligned}
      &\mathcal{G}_{O_i^{\tilde{X}}}=\left\{\mathcal{V}_{O_i^{\tilde{X}}}, \mathcal{E}_{O_i^{\tilde{X}}}\right\},  \\
      &\mathcal{V}_{O_i^{\tilde{X}}}=\left\{\tilde{x}\left(h,w\right)\ |\ \left(h,w\right)\in\mathrm{\Omega}(i)\right\},  \\
      &\mathcal{E}_{O_i^{\tilde{X}}}=\left\{(\tilde{x}\left(h,w\right),\tilde{x}\left(m,n\right))\ |\ (h,w),\ (m,n)\in\mathrm{\Omega}(i)\right\},
    \end{aligned}
\right.
\label{eq:11}
\end{equation}
where $\mathcal{V}_{O_i^{\tilde{X}}}$ is the set of feature vectors of pixels in $O_i^{\tilde{X}}$ and $\mathcal{E}_{O_i^{\tilde{X}}}$ is the set of edge between two pixels in $O_i^{\tilde{X}}$. Here, we assume that the local structural relationship exists between any two pixels within an object. Thus, $\mathcal{G}_{O_i^{\tilde{X}}}$ is designed as a fully connected graph, where an edge exists between any two vertices for connection. 
To quantitatively measure the relationship between pixels within $O_i^{\tilde{X}}$, we further construct an adjacency matrix $A_{O_i^{\tilde{X}}}$ as
\begin{equation}
\begin{split}
    A_{O_i^{\tilde{X}}}=\{\mathrm{exp}(-\phi_{1} \mathrm{dist}^\mathcal{X}(\tilde{x}&\left(h,w\right),\tilde{x}\left(m,n\right))) \mid  \\
    &(h,w), (m,n)\in\mathrm{\Omega}(i)\},
\end{split}
\label{eq:12}
\end{equation}
where $\phi_{1}$ is the bandwidth parameter and $\mathrm{dist}^\mathcal{X}(\cdot,\cdot)$ means a metric measuring the distance between two pixels in modality $\mathcal{X}$. It is common to assume the Gaussian distribution for optical images \cite{Bovolo2007a}. Concerning SAR images, the logarithm transformation in Eq. (\ref{eq:8}) can bring its distribution to near Gaussian distribution \cite{Zhan2018Log}. Thus, we use Euclidean distance in this paper since it is suitable for Gaussian data. 

\par For convenience, we further denote $\mathcal{G}_{O_i^{\tilde{X}}}$ as $\mathcal{G}_{O_i^{\tilde{X}}}=\left\{\mathcal{V}_{O_i^{\tilde{X}}}, \mathcal{E}_{O_i^{\tilde{X}}},A_{O_i^{\tilde{X}}}\right\}$. We could see that the structural information including spectral/polarimetric information and the relationship between pixels is contained in $\mathcal{G}_{O_i^{\tilde{X}}}$. Similarly, we can construct the structural graph $\mathcal{G}_{O_i^{\tilde{Y}}}=\left\{\mathcal{V}_{O_i^{\tilde{Y}}}, \mathcal{E}_{O_i^{\tilde{Y}}}, A_{O_i^{\tilde{Y}}}\right\}$ to represent the structural information contained in $O_i^{\tilde{Y}}$.

\subsection{Learning Structural Relationship Graph Representations}
\par After constructing structural graphs for pre-change and post-change images, we can exploit the two structural relationships for change detection. A feasible way is to refer to the processing flows in these similarity measure-based methods \cite{Mignotte2020,Sun2021a}. However, these methods only use the low-level information of images and the structural information contained in the graphs is not fully explored, which is inadequate to cope with the complex conditions in multimodal change detection. Therefore, we present a structural relationship graph convolutional autoencoder (SR-GCAE) to fully learn robust and representative information from structural graphs, as shown in Fig. \ref{fig:SRGCAE}. 

\par Given a structural graph $\mathcal{G}_{O_i^{\tilde{X}}}$ of pre-change image, SR-GCAE learns the structural information in $\mathcal{G}_{O_i^{\tilde{X}}}$ with several stacked graph convolutional layers as
\begin{equation}
\begin{split}
   F^{(L)}_{O_i^{\tilde{X}}}&=\mathcal{F}(\mathcal{G}_{O_i^{\tilde{X}}}) \\
   &={GCL}^{\left(L\right)}({GCL}^{\left(L-1\right)}...({GCL}^{\left(1\right)}(\mathcal{G}_{O_i^{\tilde{X}}}))...), 
\end{split}
\label{eq:13}
\end{equation}
where $L$ is the number of graph convolutional layers; $F^{(L)}_{O_i^{\tilde{X}}}\in \mathbb{R}^{N_{\Omega_i}\times C_{F}}$ is the deep graph representations with $C_{F}$ channels learned by graph convolutional layers; ${GCL}^{\left(l\right)}$ is the $l$-th graph convolutional layer, learning representations following the rule: 
\begin{equation}
F_{O_i^{\tilde{X}}}^{\left(l+1\right)}=\sigma\left({\tilde{D}_{O_i^{\tilde{X}}}}^{-\frac{1}{2}}\tilde{A}_{O_i^{\tilde{X}}}{\tilde{D}_{O_i^{\tilde{X}}}}^{-\frac{1}{2}}F_{O_i^{\tilde{X}}}^{\left(l\right)}W^{\left(l\right)}\right).
\label{eq:14}
\end{equation}

\par After learning $F^{(L)}_{O_i^{\tilde{X}}}$, the decoder of SR-GCAE reconstruct the information of $\mathcal{G}_{O_i^{\tilde{X}}}$ from $F^{(L)}_{O_i^{\tilde{X}}}$. The reconstruction process can be formulated as
\begin{equation}
{\hat{\mathcal{G}^\prime}}_{O_i^{\tilde{X}}}=\mathcal{F}^{-1}\left(F^{(L)}_{O_i^{\tilde{X}}}\right),
\label{eq:15}
\end{equation}
where ${\hat{\mathcal{G}^\prime}}_{O_i^{\tilde{X}}}$ is the reconstructed information of $\mathcal{G}_{O_i^{\tilde{X}}}$. 

\par Similarly, for post-change image $Y$, we have the above process as
\begin{equation}
{\hat{\mathcal{G}^\prime}}_{O_i^Y}=\mathcal{F}^{-1}\left(\mathcal{F}\left(\mathcal{G}_{O_i^{\tilde{Y}}}\right)\right).
\label{eq:16}
\end{equation}

\par To make $F^{(L)}_{O_i^{\tilde{X}}}$ applicable for change detection, the reconstruction way $\mathcal{F}^{-1}$ and the optimization objective $\mathcal{L}$ are the key parts. Here, we design them according to the two kinds of structural relationships in Section II-A.

\par The nonlocal structural relationship is the relationship between areas within an image. Therefore, we need to learn representative semantic features for each area. This aim can be achieved by reconstructing the vertex information:
\begin{equation}
\left\{
    \begin{aligned}
      &{\hat{\mathcal{V}}}_{O_i^{\tilde{X}}}=\mathcal{F}^{-1}\left(F^{(L)}_{O_i^{\tilde{X}}}\right)=GCL^{\left(L+1\right)}\left(F^{(L)}_{O_i^{\tilde{X}}}\right), \\
      &\mathcal{L}_{ver\ }={\frac{1}{N_{cs}}\sum_{i=1}^{N_{cs}}\frac{1}{N_{\mathrm{\Omega}_i}}}\sum_{n=1}^{N_{\mathrm{\Omega}_i}}\left({\hat{\mathcal{V}}}_{O_i^{\tilde{X}}}(n)-\mathcal{V}_{O_i^{\tilde{X}}}(n)\right)^2,
    \end{aligned}
\right.
\label{eq:17}
\end{equation}
where ${\hat{\mathcal{V}}}_{O_i^{\tilde{X}}}$ is the reconstructed vertex information of $\mathcal{G}_{O_i^{\tilde{X}}}$ and $\mathcal{L}_{ver}$ is the optimization objective for reconstructing vertex information. Through optimizing $\mathcal{L}_{ver\ }$, the vertex information can be encoded in $F^{(L)}_{O_i^{\tilde{X}}}$. We denote it as $F_{O_i^{\tilde{X}}}^{ver}$. Note that different from the symmetric structure commonly seen in the standard autoencoder and autoencoder-based work \cite{Wu2021b,Liu2019c,Mou2018}, we only adopt a simple decoder with one graph convolutional layer to reconstruct the vertex information. An intuitive idea for this design is that the weak reconstruction ability of the decoder can enforce the encoder to learn more representative semantic features.

\par The local structural relationship is the relationship between pixels within an area. Thus, we reconstruct the edge information of $\mathcal{G}_{O_i^{\tilde{X}}}$ from $F^{(L)}_{O_i^{\tilde{X}}}$ so that the relationship between pixels can be represented by $F^{(L)}_{O_i^{\tilde{X}}}$. However, it is difficult to explicitly reconstruct the edge information in the way of graph convolutional layers like the vertex information reconstruction in Eq. (\ref{eq:17}). Therefore, we reconstruct the edge information from $F^{(L)}_{O_i^{\tilde{X}}}$ directly as follows: 
\begin{equation}
\left\{
    \begin{aligned}
      &{\hat{A}}_{O_i^{\tilde{X}}}=\mathcal{F}^{-1}\left(F^{(L)}_{O_i^{\tilde{X}}}\right)=\sigma\left(F^{(L)}_{O_i^{\tilde{X}}}(F^{(L)}_{O_i^{\tilde{X}}})^T\right),  \\
      &\mathcal{L}_{eg}=\frac{1}{N_{cs}}\sum_{i=1}^{N_{cs}}\frac{1}{N_{\mathrm{\Omega}_i}^2}\sum_{m=1}^{N_{\mathrm{\Omega}_i}}\sum_{n=1}^{N_{\mathrm{\Omega}_i}}\left({\hat{A}}_{O_i^{\tilde{X}}}\left(m,n\right)-{A}_{O_i^{\tilde{X}}}\left(m,n\right)\right)^2,
    \end{aligned}
\right.
\label{eq:18}
\end{equation}
where ${\hat{A}}_{O_i^{\tilde{X}}} \in \mathbb{R}^{N_{\Omega_{i}}\times N_{\Omega_{i}}}$ is the reconstructed adjacent matrix from $F^{(L)}_{O_i^{\tilde{X}}}$, $\sigma$ is an activation function, and $\mathcal{L}_{eg}$ is the optimization objective for reconstructing edge information. Through optimizing $\mathcal{L}_{eg}$, the edge information can be encoded in $F^{(L)}_{O_i^{\tilde{X}}}$. We denote it as $F_{O_i^{\tilde{X}}}^{eg}$. 

\par Similarly, SR-GCAE learns $F_{O_i^{\tilde{Y}}}^{eg}$ and $F_{O_i^{\tilde{Y}}}^{ver}$ from $\mathcal{G}_{O_i^{\tilde{Y}}}$ for post-change image $Y$.

\subsection{Change Information Mapping}
\par After learning $F_{O_i^{\tilde{X}}}^{eg}$, $F_{O_i^{\tilde{X}}}^{ver}$, $F_{O_i^{\tilde{Y}}}^{eg}$, and $F_{O_i^{\tilde{Y}}}^{ver}$ from $X$ and $Y$, we can perform change detection using the rich robust structural information contained in these features. 

\par If change event happens in the area of $\Omega_i$, the structural relationship between pixels in $\mathrm{\Omega}_i$ cannot preserve consistent in $O_i^{\tilde{X}}$ and $O_i^{\tilde{Y}}$. Thus, this change can be reflected in $F_{O_i^{\tilde{X}}}^{eg}$ and $F_{O_i^{\tilde{Y}}}^{eg}$. Therefore, the intuitive and simple idea is to calculate the distance between $F_{O_i^{\tilde{X}}}^{eg}$ and $F_{O_i^{\tilde{Y}}}^{eg}$:
\begin{equation}
d_{\mathrm{\Omega}_i}^{lcl}=\mathrm{dist} (F_{O_i^{\tilde{X}}}^{eg}, F_{O_i^{\tilde{Y}}}^{eg}),
\label{eq:19}
\end{equation}
where $\mathrm{dist} (F_{O_i^{\tilde{X}}}^{eg},F_{O_i^{\tilde{Y}}}^{eg})$ is the distance between deep edge representations $F_{O_i^{\tilde{X}}}^{eg}$ and $F_{O_i^{\tilde{Y}}}^{eg}$. We apply $L_{1}$-distance here, i.e., 
\begin{equation}
\mathrm{dist}\left(F_{O_i^{\tilde{X}}}^{eg},F_{O_i^{\tilde{Y}}}^{eg}\right)=\frac{1}{N_{\mathrm{\Omega}_i}}\sum_{j=1}^{N_{\mathrm{\Omega}_i}} \left| F_{O_i^{\tilde{X}}}^{eg}(j)-F_{O_i^{\tilde{Y}}}^{eg}\left(j\right)\right|.
\label{eq:19_2}
\end{equation}

\par The local difference image $\mathcal{DI}^{lcl}\in \mathbb{R}^{H\times W}$ can be obtained by assigning $d_{\mathrm{\Omega}_i}^{lcl}$ to the specific pixels according to the segmentation map $\mathrm{\Omega}$:
\begin{equation}
\mathcal{DI}^{lcl}\left(h,w\right)=d_{\mathrm{\Omega}_i}^{lcl},
\label{eq:20}
\end{equation}
where $\left(h, w\right)\ \in\mathrm{\Omega}_i,  i=1,2,\ldots,N_{cs}$. 

\par If change event happens in the area of $\mathrm{\Omega}_i$, the structural relationship between $\mathrm{\Omega}_i$ and its similar objects is not consistent in $X$ and $Y$. Accordingly, we further construct a nonlocal structural graph for $O_i^{\tilde{X}}$ by finding its most similar $K$ objects and calculating their similarities based on deep graph representations to represent the nonlocal structural relationship as

\begin{small}
\begin{equation}
\left\{
    \begin{aligned}
      &\mathcal{G}_{O_i^{\tilde{X}}}^\mathcal{X}=\left\{\mathcal{V}_{O_i^{\tilde{X}}}^{\mathcal{X}},\mathcal{E}_{O_i^{\tilde{X}}}^{\mathcal{X}},A_{O_i^{\tilde{X}}}^{\mathcal{X}}\right\}, \\
      &\mathcal{V}_{O_i^{\tilde{X}}}^\mathcal{X}=\left\{F_{O_k^{\tilde{X}}}^{ver},k=1,2,...,K\right\}, \\
      &\mathcal{E}_{O_i^{\tilde{X}}}^\mathcal{X}=\left\{\left(F_{O_i^{\tilde{X}}}^{ver},F_{O_k^{\tilde{X}}}^{ver}\right)\ |\ F_{O_k^{\tilde{X}}}^{ver}\in \mathcal{V}_{O_i^{\tilde{X}}}^\mathcal{X}\right\}, \\
      &A_{O_i^{\tilde{X}}}^\mathcal{X}=\left\{exp(-\phi\ dist^\mathcal{X}(F_{O_i^{\tilde{X}}}^{ver},F_{O_k^{\tilde{X}}}^{ver})) \mid (F_{O_i^{\tilde{X}}}^{ver},F_{O_k^{\tilde{X}}}^{ver})\in \mathcal{E}_{O_i^{\tilde{X}}}^\mathcal{X}\right\},
    \end{aligned}
\right.
\label{eq:21}
\end{equation}
\end{small}
where $\mathcal{G}_{O_i^{\tilde{X}}}^\mathcal{X}$ is the nonlocal structural graph constructed for $O_i^{\tilde{X}}$, $O_k^{\tilde{X}}$ is the $k$-th object that is most similar to $O_i^{\tilde{X}}$. 

\par In post-change image $Y$, we can also construct such a nonlocal graph $\mathcal{G}_{O_i^{\tilde{Y}}}^\mathcal{X}=\{\mathcal{V}_{O_i^{\tilde{Y}}}^\mathcal{X},\mathcal{E}_{O_i^{\tilde{Y}}}^\mathcal{X},A_{O_i^{\tilde{Y}}}^\mathcal{X}\}$ based on the nonlocal structural relationship found for $O_i^{\tilde{X}}$. Then, we can compare the difference between $\mathcal{G}_{O_i^{\tilde{X}}}^\mathcal{X}$ and $\mathcal{G}_{O_i^{\tilde{Y}}}^\mathcal{X}$ to determine whether change event happens in the area $\mathrm{\Omega}_i$ or not: 
\begin{small}
\begin{equation}
\begin{split}
d_{\mathrm{\Omega}_i}^{\mathcal{X}-\mathcal{Y}}=\frac{1}{K}\sum_{k=1}^{K}\sum_{c=1}^{C_{F}} | &\mathrm{exp}(-\phi_{2} \mathrm{dist}^\mathcal{X}\left({||F}_{O_i^{\tilde{X}}}^{ver}(c)||_p,{||F}_{O_k^{\tilde{X}}}^{ver}(c)||_p\right) \\
-&\mathrm{exp}(-\phi_{2}\mathrm{dist}^\mathcal{X}\left({||F}_{O_i^{\tilde{Y}}}^{ver}(c)||_p,{||F}_{O_k^{\tilde{Y}}}^{ver}(c)||_p\right) |
\end{split}
\label{eq:22}
\end{equation}
\end{small}
Here, since the dimension of $F_{O_i^{\tilde{X}}}^{ver}$ and $F_{O_k^{\tilde{X}}}^{ver}$ could be different, we cannot calculate the distance between them directly. To overcome this problem, the $p$-norm is first employed on each channel of $F_{O_i^{\tilde{X}}}^{ver}$ and $F_{O_k^{\tilde{X}}}^{ver}$ before calculating $dist^\mathcal{X}(\cdot,\cdot)$. The rationale for this operation is that the message passing step of graph convolution enforces neighboring vertices to get similar representations \cite{kipf2016semi,wu2019simplifying}. Common values of $p$ are one, two, and infinity. We use the $L_{1}$ norm in this paper. 

\par The above step can also be executed by finding the most similar objects for $O_i^{\tilde{Y}}$ and mapping this relationship to $X$:
\begin{small}
\begin{equation}
\begin{split}
d_{\mathrm{\Omega}_i}^{\mathcal{Y}-\mathcal{X}}=\frac{1}{K}\sum_{k=1}^{K}\sum_{c=1}^{C_{F}} | &\mathrm{exp}(-\phi_{2} \mathrm{dist}^\mathcal{Y}\left({||F}_{O_i^{\tilde{Y}}}^{ver}(c)||_p,{||F}_{O_k^{\tilde{Y}}}^{ver}(c)||_p\right) \\
-&\mathrm{exp}(-\phi_{2}\mathrm{dist}^\mathcal{Y}\left({||F}_{O_i^{\tilde{X}}}^{ver}(c)||_p,{||F}_{O_k^{\tilde{x}}}^{ver}(c)||_p\right) |
\end{split}
\label{eq:23}
\end{equation}
\end{small}
\par After these two steps, the nonlocal difference image can be denoted as
\begin{equation}
\mathcal{DI}^{nlcl}\left(h,w\right)=d_{\mathrm{\Omega}_i}^{\mathcal{X}-\mathcal{Y}}+d_{\mathrm{\Omega}_i}^{\mathcal{Y}-\mathcal{X}},
\label{eq:24}
\end{equation}
where $\left(h, w\right)\ \in\mathrm{\Omega}_i, i=1,2,\ldots,N_{cs}$.
\par Once $\mathcal{DI}^{lcl}$ and $\mathcal{D}\mathcal{I}^{nlcl}$ are obtained, we can fuse them to get a more robust difference image. Compared to simply adding two difference maps together, it is more appropriate to let the difference map with higher quality take a greater weight in the fusion process. In this paper, we present an effective adaptive fusion strategy to fuse $\mathcal{D}\mathcal{I}^{lcl}$ and $\mathcal{D}\mathcal{I}^{nlcl}$:
\begin{equation}
\begin{split}
  \mathcal{D}\mathcal{I}^{final}&=Fuse\left(\mathcal{D}\mathcal{I}^{lcl},\mathcal{D}\mathcal{I}^{nlcl}\right) \\
  &=\frac{\mathcal{V}\left(\mathcal{D}\mathcal{I}^{lcl}\right)\mathcal{D}\mathcal{I}^{lcl}+\mathcal{V}\left(\mathcal{D}\mathcal{I}^{nlcl}\right)\mathcal{D}\mathcal{I}^{nlcl}}{\mathcal{V}\left(\mathcal{D}\mathcal{I}^{lcl}\right)+\mathcal{V}\left(\mathcal{D}\mathcal{I}^{nlcl}\right)},  
\end{split}
\label{eq:25}
\end{equation}
where $\mathcal{D}\mathcal{I}^{final}$ is the final difference map and $\mathcal{V}\left(\mathcal{DI}\right)$ is the variance of change intensity in $\mathcal{DI}$ calculated as 
\begin{equation}
\mathcal{V}\left(\mathcal{DI}\right)=\frac{1}{HW}\sum_{h=1}^{H}\sum_{w=1}^{W}(\mathcal{DI}(h,w)-\frac{1}{HW}\sum_{h=1}^{H}\sum_{w=1}^{W}{\mathcal{DI}(h,w))}^2
\end{equation}
\par The idea of this adaptive fusion strategy is very simple and straightforward: changed and unchanged pixels are very discriminative in the difference image with high quality. This high discrimination would be reflected in the variance of the change intensity of the difference image.

\subsection{Morphological Post-processing}
After $\mathcal{D}\mathcal{I}^{final}$ is generated, the change detection problem can be treated as a binary classification problem. The threshold segmentation methods can be performed on $\mathcal{DI}^{final}$ to map pixels to change class $\omega_{c}$ and non-change class $\omega_{nc}$:
\begin{equation}
\mathcal{CM}(h, w)=\left\{\begin{array}{c}
\omega_{n c}, \mathcal{DI}^{final}(h, w) \leq T, \\
\omega_{c}, \mathcal{DI}^{final}(h, w)>T,
\end{array}\right.
\label{eq:26}
\end{equation}
where $\mathcal{CM}$ is the binary change map and $T$ is the threshold obtained by threshold segmentation methods, such as Otsu's method \cite{otsu1979} and expectation maximization \cite{moon1996}. 

\begin{figure}[!t]
  \centering

  \subfloat[]{
    \includegraphics[width=1.05in]{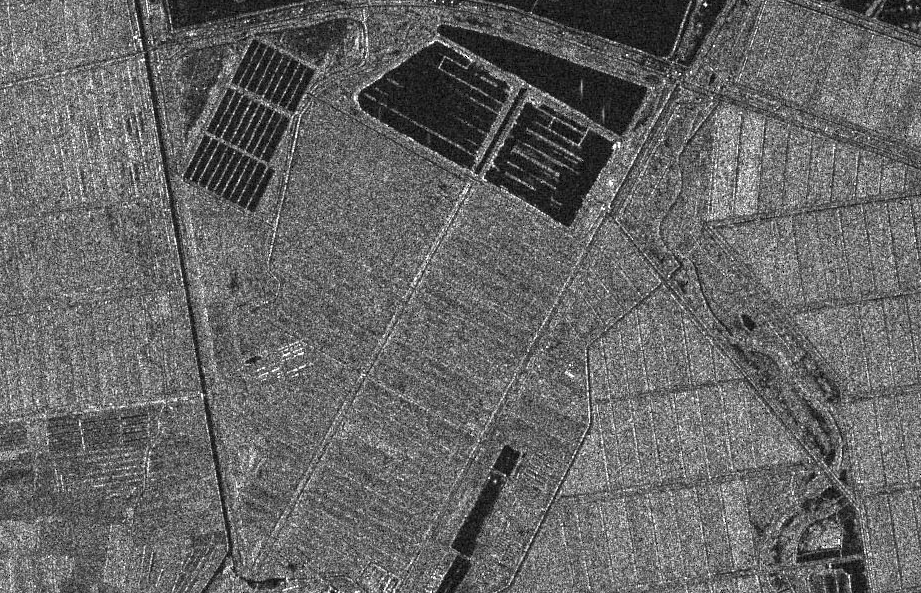}
  \label{fig_first_case}}
  \hfil
  \subfloat[]{
    \includegraphics[width=1.05in]{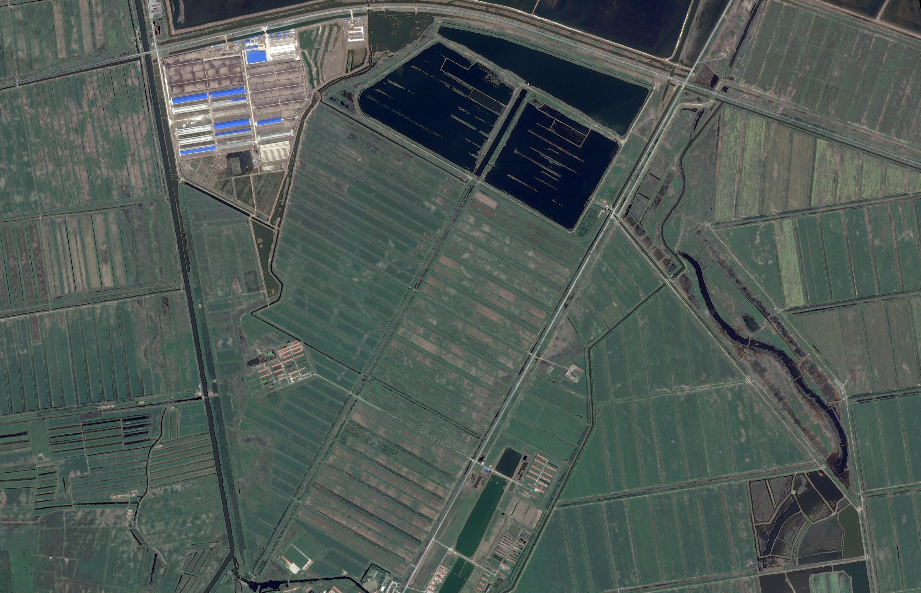}
  \label{fig_second_case}}
  \hfil
  \subfloat[]{
    \includegraphics[width=1.05in]{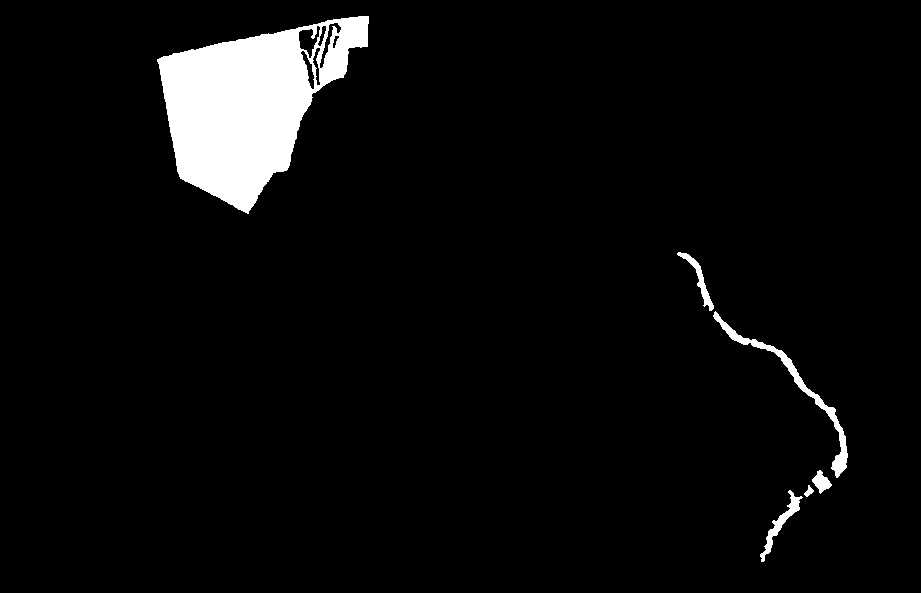}
  \label{fig_third_case}}

  \caption{Shuguang dataset. (a) Pre-change SAR image. (b) Post-change optical image. (c) Reference map.}
  \label{SG_dataset}
\end{figure}

\begin{figure}[!t]
  \centering
  \subfloat[]{
    \includegraphics[width=1.05in]{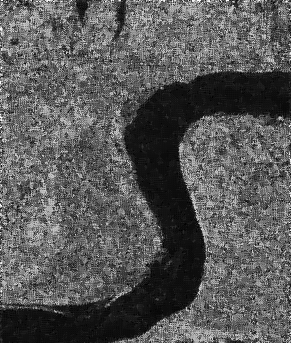}
  \label{fig_first_case}}
  \hfil
  \subfloat[]{
    \includegraphics[width=1.05in]{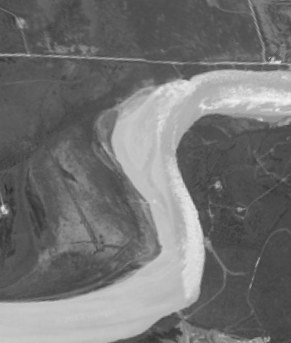}
  \label{fig_second_case}}
  \hfil
  \subfloat[]{
    \includegraphics[width=1.05in]{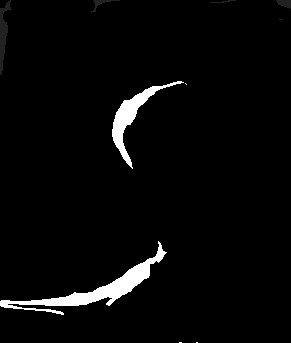}
  \label{fig_third_case}}

  \caption{River dataset. (a) Pre-change SAR image. (b) Post-change panchromatic image. (c) Reference map. }
  \label{YR_dataset}
\end{figure}

\begin{figure}[!t]
  \centering
  \subfloat[]{
    \includegraphics[width=1.05in]{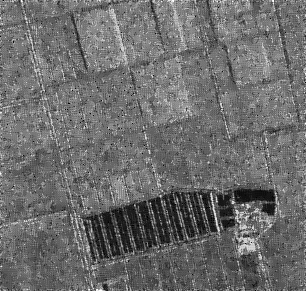}
  \label{fig_first_case}}
  \hfil
  \subfloat[]{
    \includegraphics[width=1.05in]{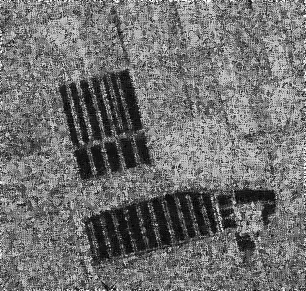}
  \label{fig_second_case}}
  \hfil
  \subfloat[]{
    \includegraphics[width=1.05in]{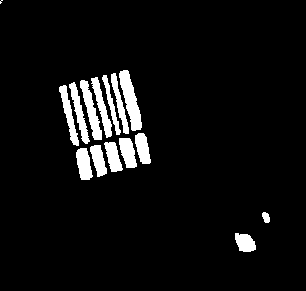}
  \label{fig_third_case}}

  \caption{Farmland dataset. (a) Pre-change image (SAR image with one look). (b) Post-change image (SAR image with four looks). (c) Reference map.}
  \label{FL_dataset}
\end{figure}
\par In addition, the difficult conditions of unsupervised multimodal change detection often cause inaccurate detection pixels. Therefore, applying some post-processing methods to refine the detection results is necessary. Some approaches choose probability graph models for post-processing \cite{Touati2020a,Luppino2019Unsupervised,Sun2021a}. However, the process of optimizing such models is often complex and computationally intensive. To keep our framework simple and practical, the morphological filtering is applied as our post-processing method.
\par Specifically, a close operation is first performed to fill possible voids within changed areas:
\begin{equation}
\widehat{\mathcal{CM}}=\mathcal{CM}\bullet\mathcal{K}_c=\left(\mathcal{CM}\oplus\mathcal{K}_c\right)\ominus\mathcal{K}_c\ ,
\label{eq:27}
\end{equation}
where $\bullet$ is the open operator composed of a dilation operator $\oplus$ and an erosion operator $\ominus$, $\mathcal{K}_{c}$ is the structuring element (filtering kernel) of the close operator. 

\begin{algorithm}[!t]
  \algsetup{linenosize=\small} \small
	\caption{Process of unsupervised multimodal change detection via SR-GCAE.}
	\label{alg:SR-GCAE}
	\begin{algorithmic}[1]
		\REQUIRE~~\\
		 Pre-change image $X$ and Post-change $Y$ with different modalities;\\
		\ENSURE ~~\\
		The difference image $\mathcal{D}\mathcal{I}^{final}$ and binary change map $\widetilde{\mathcal{CM}}$;\\
	\STATE Align $X$ and $Y$ geometrically;
    \STATE Normalize $X$ and $Y$ according to their modality;
    \STATE Construct structural graphs $G_{O_i^{\tilde{X}}}$ and $G_{O_i^{\tilde{Y}}}$;
    \STATE Learn graph representations $F_{O_i^{\tilde{X}}}^{eg}$ and $F_{O_i^{\tilde{Y}}}^{eg}$ via SR-GCAE with the optimization objective of edge reconstruction: $$\mathcal{L}_{eg}=\frac{1}{N_{cs}}\sum_{i=1}^{N_{cs}}\frac{1}{N_{\mathrm{\Omega}_i}^2}\sum_{m=1}^{N_{\mathrm{\Omega}_i}}\sum_{n=1}^{N_{\mathrm{\Omega}_i}}\left({\hat{A}}_{O_i^{\tilde{X}}}\left(m,n\right)-A_{O_i^{\tilde{X}}}\left(m,n\right)\right)^2;$$
    \STATE Learn graph representations $F_{O_i^{\tilde{X}}}^{ver}$ and $F_{O_i^{\tilde{Y}}}^{ver}$ via SR-GCAE with the optimization objective of vertex reconstruction:
    $$\mathcal{L}_{ver\ }={\frac{1}{N_{cs}}\sum_{i=1}^{N_{cs}}\frac{1}{N_{\mathrm{\Omega}_i}}}\sum_{n=1}^{N_{\mathrm{\Omega}_i}}({\hat{\mathcal{V}}}_{O_i^{\tilde{X}}}(n)-\mathcal{V}_{O_i^{\tilde{X}}}(n))^2;$$
    \STATE Calculate local similarity difference image $\mathcal{D}\mathcal{I}^{lcl}$ using $F_{O_i^{\tilde{X}}}^{eg}$ and $F_{O_i^{\tilde{Y}}}^{eg}$;
    \STATE Calculate nonlocal similarity difference image $\mathcal{D}\mathcal{I}^{nlcl}$ using $F_{O_i^{\tilde{X}}}^{ver}$ and $F_{O_i^{\tilde{Y}}}^{ver}$;
    \STATE Fuse local and nonlocal difference images to get the final difference image:
    $$\mathcal{D}\mathcal{I}^{final}=Fuse\left(\mathcal{D}\mathcal{I}^{lcl},\mathcal{D}\mathcal{I}^{nlcl}\right);$$
    \STATE Perform threshold segmentation method to get the binary change map $\mathcal{CM}$; 
    \STATE Perform morphological filtering to refine the change map:$$\widetilde{\mathcal{CM}}=MF(\mathcal{CM},\mathcal{K}_c,\mathcal{K}_o);$$
    \RETURN $\mathcal{DI}^{final}$ and $\tilde{\mathcal{CM}}$;
	\end{algorithmic}
\end{algorithm}

\par Then, an open operation is performed to erase those isolated changed pixels: 
\begin{equation}
\widetilde{\mathcal{CM}}=\mathcal{CM}\circ\mathcal{K}_o=\left(\widehat{\mathcal{CM}}\ominus\mathcal{K}_o\right)\oplus\mathcal{K}_o,
\label{eq:28}
\end{equation}
where $\mathcal{K}_{o}$ is the structuring element of the open operator and $\widetilde{\mathcal{CM}}$ is the final refined change map. 
\par The above process can be formed as
\begin{equation}
\widetilde{\mathcal{CM}}=MF(\mathcal{CM},\mathcal{K}_c,\mathcal{K}_o)=(\mathcal{CM}\bullet\mathcal{K}_c)\circ\mathcal{K}_o.
\label{eq:29}
\end{equation}

\par Summarizing all of the aforementioned contents, the overall framework of our SR-GCAE is elaborated in Algorithm \ref{alg:SR-GCAE}.

\begin{table*}[t]
  \setlength\tabcolsep{5pt}
  \renewcommand{\arraystretch}{1.4}
  \caption{Information of five multimodal change detection datasets}
  \label{dataset_info}
  \centering
  \begin{tabular}{c c c c c}
    \hline			
    \textbf{Dataset}	& \textbf{Sensor} &	\textbf{Size} & \textbf{Location} & \textbf{Change Event} \\
    \hline\hline
    Shuguang &	Radarsat-2/Google Earth &	593$\times$921$\times$1/3	& Dongying, China & Constructions\\
    River	& Radarsat-2/Landsat-7 &	291$\times$343$\times$1/1	& Yellow River, China &	River flood\\
    Farmland &	Radarsat-2 (single/four look) &	306$\times$291$\times$1/1 &	Eastern China &	Farmland reclaim \\
    Texas &	Landsat-5/EO-1 ALI	& 1534$\times$808$\times$7/10&	Texas, USA&	Forest fire\\
    Hanyang	&GaoFen-2/Gaofen-2	&1000$\times$1000$\times$4/4&	Wuhan, China&	Urban constructions / water bloom\\
    \hline
  \end{tabular}
\end{table*}

\section{Experiment}\label{sec:4}
\subsection{Data Description}
\par To verify the effectiveness of our method, four heterogeneous and one homogeneous change detection datasets are used in the experimental part. 

\par The first dataset is the Shuguang dataset, consisting of a SAR image and an optical aerial image. Fig. \ref{SG_dataset} (a)–(c) shows the two images and the reference map, respectively. The preprocessed SAR and optical images have the size of 921$\times$593 pixels and were acquired in 2008 and 2012, respectively, covering a part of a village in Shandong Province, China. 
\begin{figure}[!t]
  \centering
  \subfloat[]{
    \includegraphics[width=1.05in]{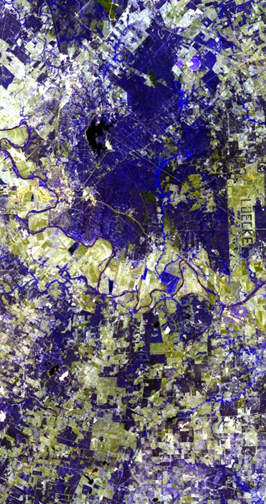}
  \label{fig_first_case}}
  \hfil
  \subfloat[]{
    \includegraphics[width=1.05in]{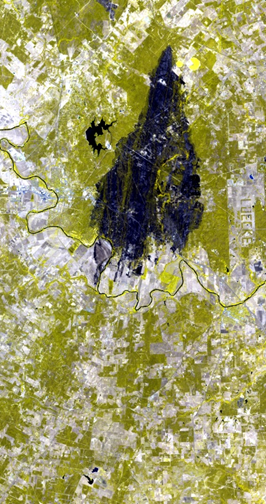}
  \label{fig_second_case}}
  \hfil
  \subfloat[]{
    \includegraphics[width=1.05in]{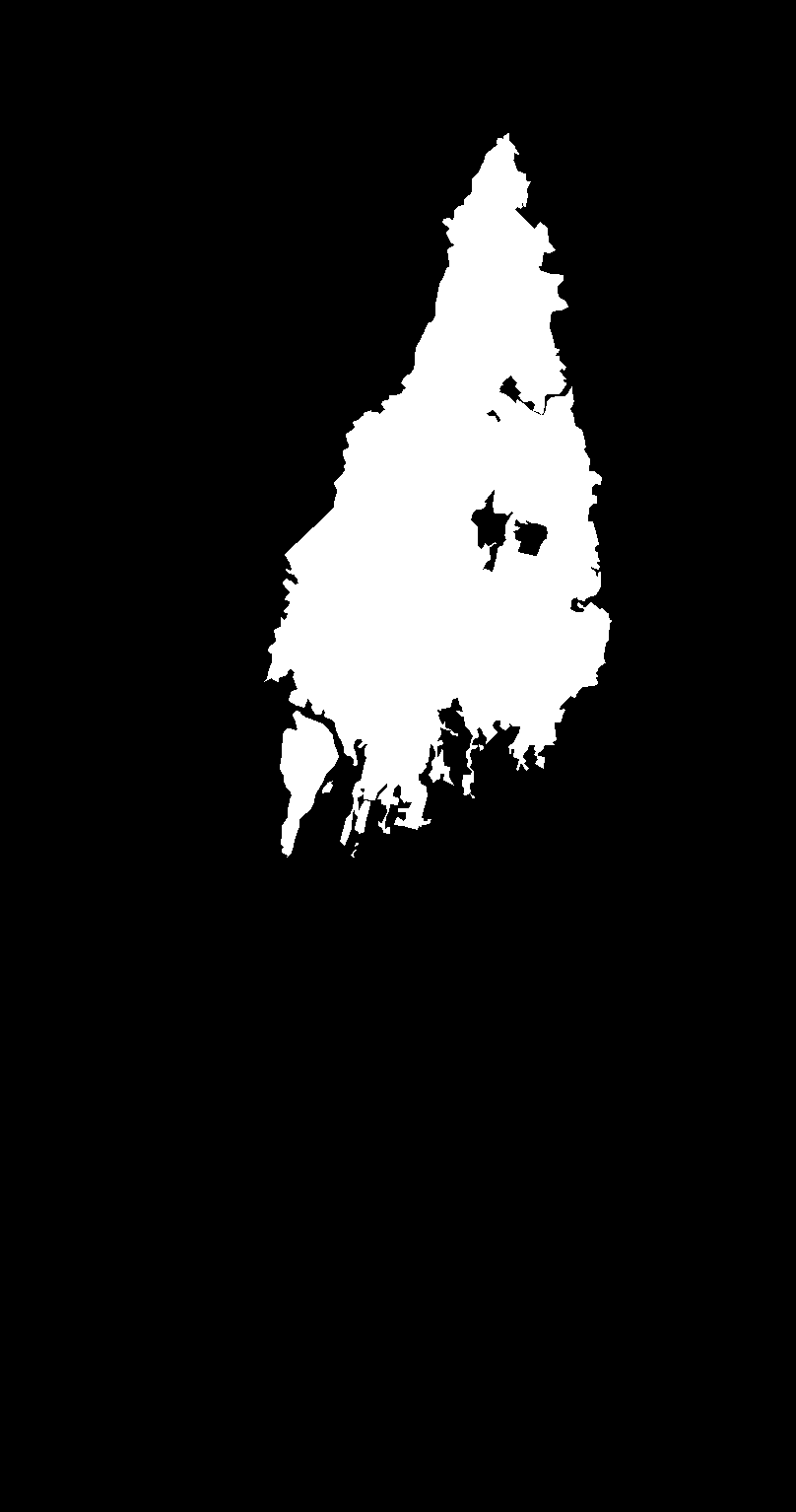}
  \label{fig_third_case}}

  \caption{Texas dataset. (a) Pre-change multispectral image. (b) Post-change multi-spectral image. (c) Reference map.}
  \label{Texas_dataset}
\end{figure}

\begin{figure}[!t]
  \centering
  \subfloat[]{
    \includegraphics[width=1.05in]{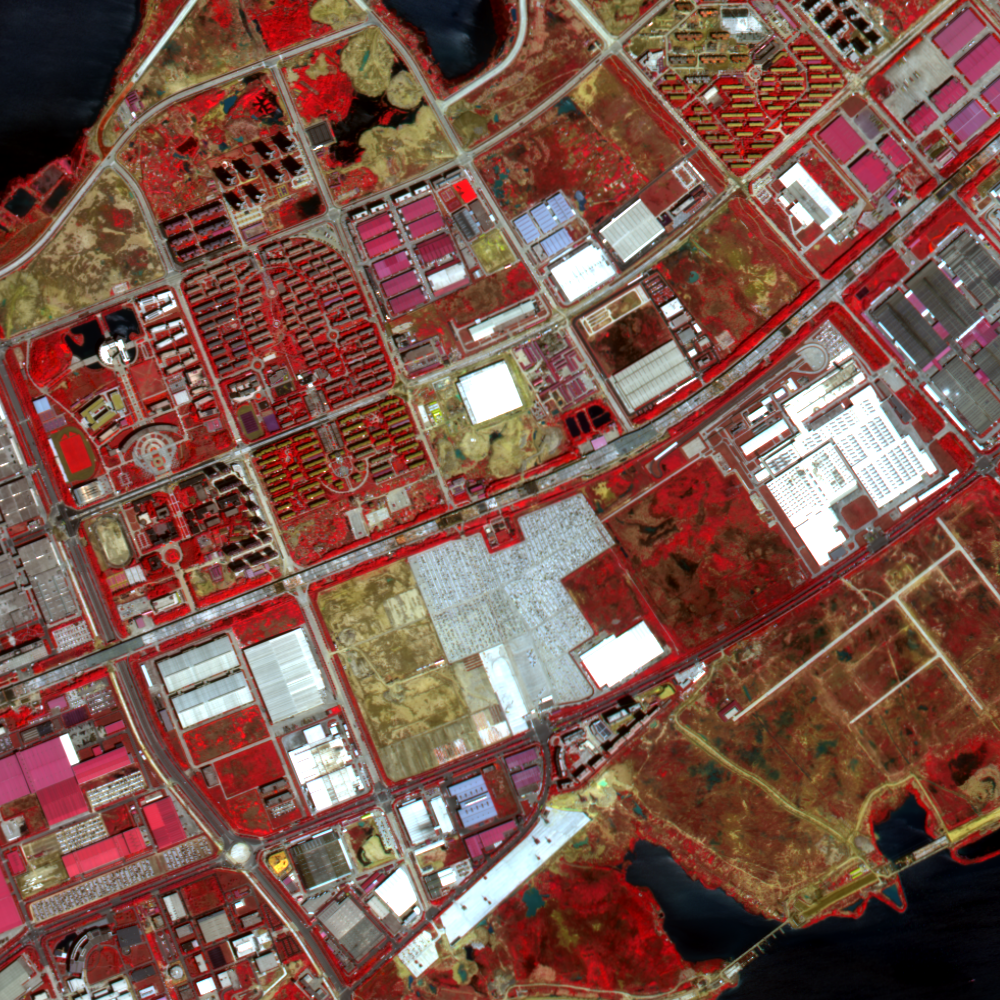}
  \label{fig_first_case}}
  \hfil
  \subfloat[]{
    \includegraphics[width=1.05in]{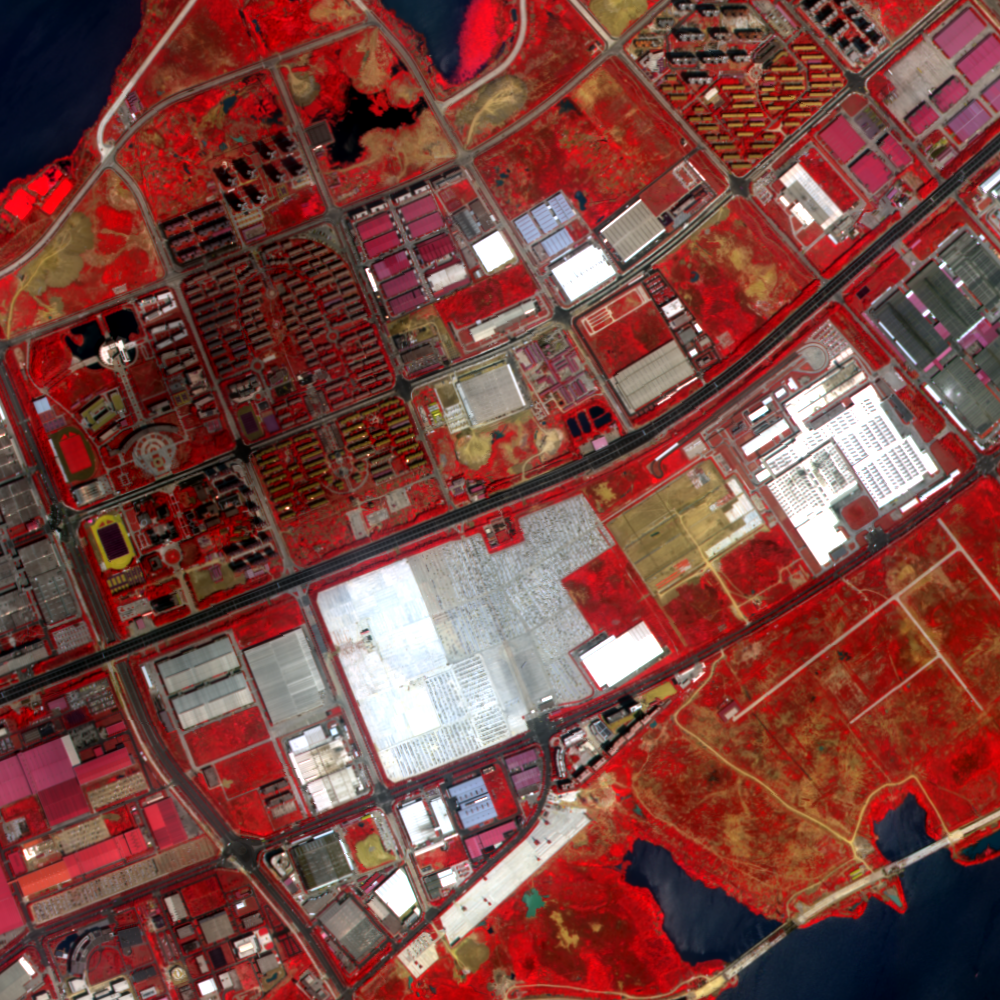}
  \label{fig_second_case}}
  \hfil
  \subfloat[]{
    \includegraphics[width=1.05in]{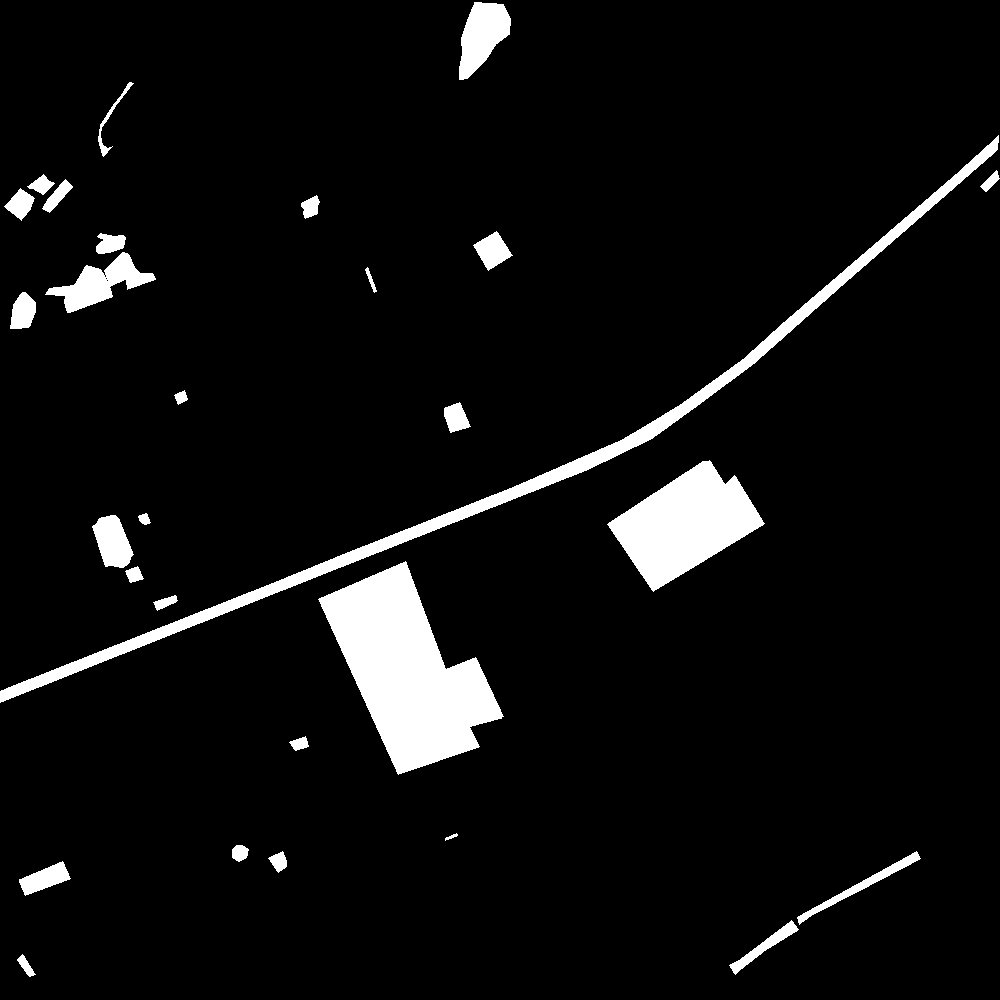}
  \label{fig_third_case}}

  \caption{Hanyang dataset. (a) Pre-change multi-spectral image. (b) Post-change multi-spectral image. (c) Reference map.}
  \label{HY_dataset}
\end{figure}
\par The second dataset is the River dataset, consisting of a SAR image and a panchromatic image with a size of 291$\times$343 pixels. The pre-change SAR image was captured by Radarsat-2 at the Yellow River Estuary in 2008. The post-change panchromatic image was captured by Landsat-7 in September 2010. The main change event in this dataset was bank erosion caused by flooding. Fig. \ref{YR_dataset} (a)–(c) shows the two images and the reference map, respectively.

\par The third dataset called the Farmland dataset has two SAR images with the same size of 306 $\times$ 291 pixels, as shown in Fig. \ref{FL_dataset} (a)-(c). These two SAR images were captured by Radarsat-2 in 2008 and 2009, respectively, covering farmland along the Yellow River in China. Although the two images were imaged by the same sensor, the pre-change image and post-change image are single-look and four-look, respectively, thereby showing different modalities.

\begin{figure*}[!ht]
  \centering
  \subfloat[]{
    \includegraphics[height=7in]{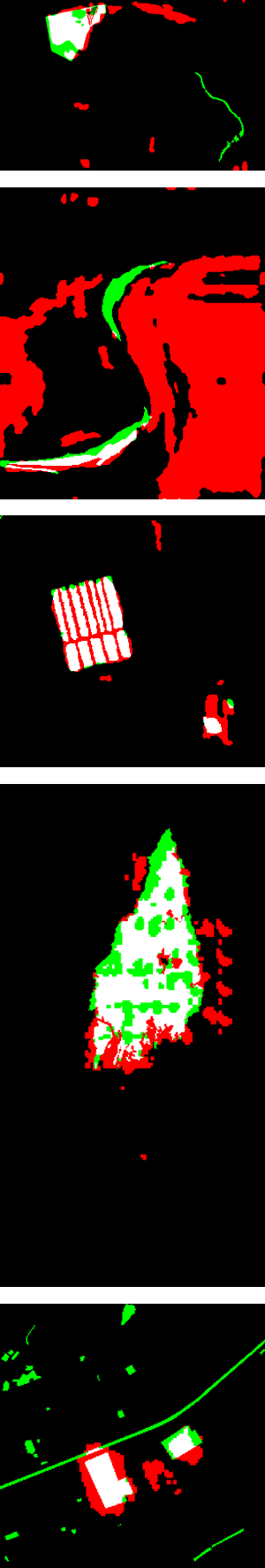}
  \label{fig_first_case}}
  \hfil
  \subfloat[]{
    \includegraphics[height=7in]{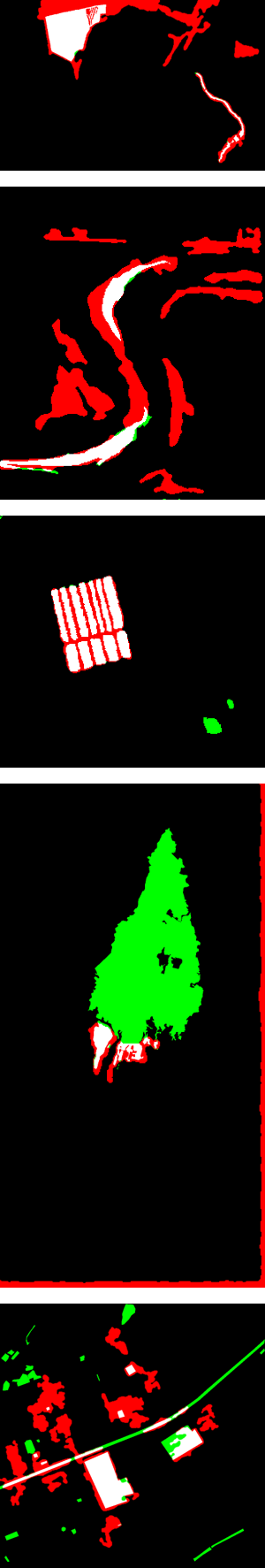}
  \label{fig_second_case}}
  \hfil
  \subfloat[]{
    \includegraphics[height=7in]{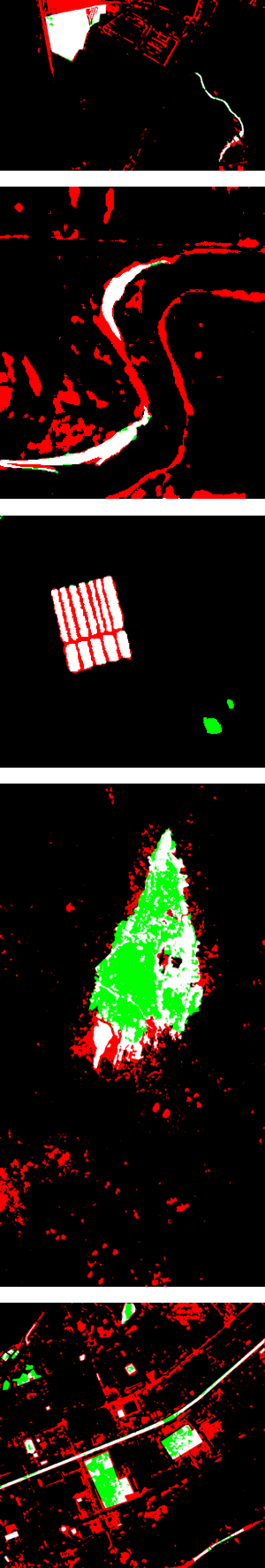}
  \label{fig_third_case}}
  \hfil
 \subfloat[]{
    \includegraphics[height=7in]{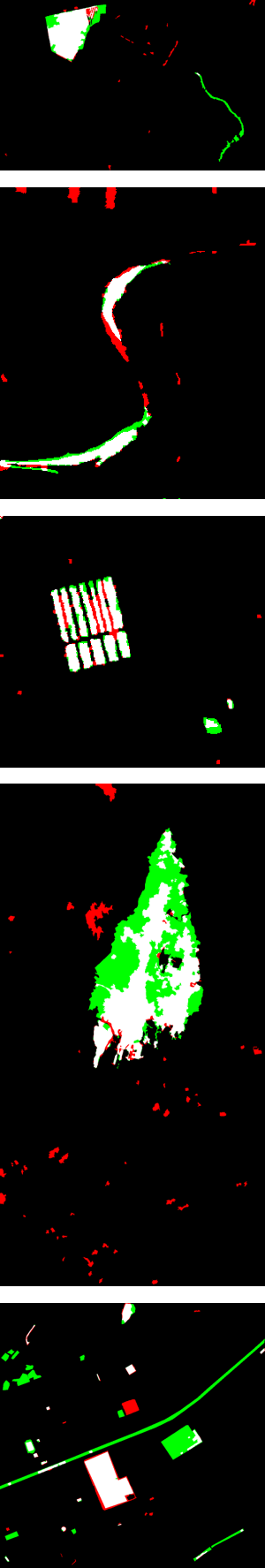}
  \label{fig_second_case}}
  \hfil
   \subfloat[]{
    \includegraphics[height=7in]{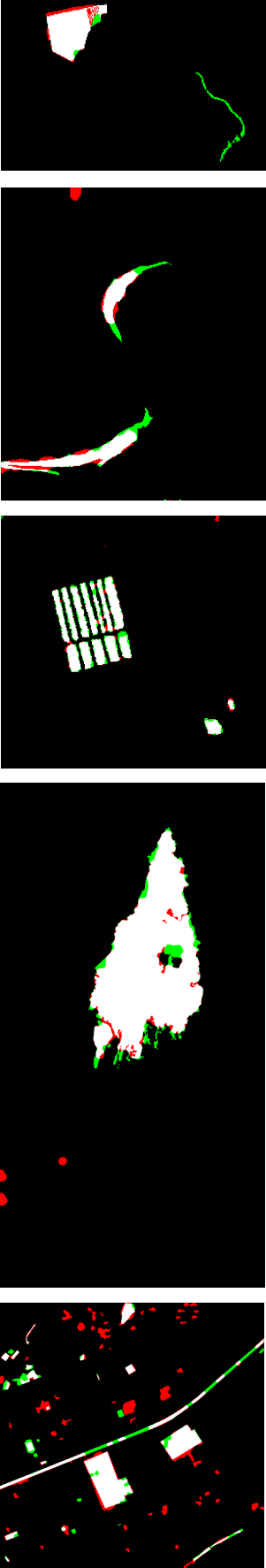}
  \label{fig_second_case}}
  \caption{Change maps obtained by different methods on the five multimodal change detection datasets. (a) M3CD. (b) FPMS. (c) NPSG. (d) IRGMcS. (e) SR-GCAE. In change maps, White: true positives (TP); Red: false positives (FP); Black: true negatives (TN); Green: false negatives (FN). }
  \label{cd_result}
\end{figure*}
\par The fourth dataset is composed of two multispectral images, as shown in Fig. \ref{Texas_dataset} (a)-(c). The pre-change image was captured by Landsat-5 with seven bands and the post-change image was captured by EO-1 ALI with 10 bands. Both images have a size of 1534$\times$808 pixels, showing the changes in a forest area in Texas States, USA, caused by wildfire. 

\par Besides, we also perform the experiments on a homogeneous dataset, the Hanyang dataset, to further prove the generalization of our method in homogeneous change detection. The two images of the Hanyang dataset consist of four spectral bands with 1000$\times$1000 pixels and they have a spatial resolution of 4m/pixel. The pseudo-color images and the reference map are shown in Fig. \ref{HY_dataset}. 

\par We could see that these five multimodal change detection datasets have different spatial and spectral resolutions, cover different modality combinations, and reflect different types of change events. It can therefore fully evaluate the effectiveness and generalizability of the proposed method in diverse conditions. The information of these five datasets is summarized in Table \ref{dataset_info}.

\begin{table*}[!ht]
  \setlength\tabcolsep{5pt}
  
  \renewcommand{\arraystretch}{1.4}
  \caption{\centering{Accuracy assessment on change maps obtained by different methods on the five multimodal change detection datasets. The highest accuracy is highlighted in bold and the second highest accuracy is underlined}}
  \label{acc_ass}
  \centering
  \begin{tabular}{c |c c c |c c c | c c c| c c c |c c c}
    \hline			
    \multirow{2}{*}{\textbf{Method}} & \multicolumn{3}{c}{\textbf{Shuguang}} & \multicolumn{3}{c}{\textbf{River}} & \multicolumn{3}{c}{\textbf{Farmland}} & 
    \multicolumn{3}{c}{\textbf{Texas}} &
    \multicolumn{3}{c}{\textbf{Hanyang}}\\
    \cline{2-16} 
    & OA & F1 & KC & OA & F1 & KC & OA & F1 & KC & OA & F1 & KC & OA & F1 & KC \\
    \hline\hline
    M3CD \cite{Touati2020a} & 95.64 & 0.5794 & 0.5569 & 71.17 & 0.1388 & 0.0865 & 95.55 & 0.7164 & 0.6938 & 92.97 & \underline{0.6644} & \underline{0.6252} & 91.05 & 0.5514 & 0.5049 \\ 	
    FPMS \cite{Mignotte2020} & 92.43 & 0.5464 & 0.5141 & 87.22 & 0.3150 & 0.2768 & 96.91 & 0.7805 & 0.7644 & 86.49 & 0.0824 & 0.0249 & 86.88 & 0.5252 & 0.4500 \\ 
    NPSG \cite{Sun2021c} & 95.02 & 0.6387 & 0.6151 & 89.11 & 0.3572 & 0.3220 & 92.45 & 0.5970 & 0.5614 & 89.93 & 0.4826 & 0.4273 & 87.94 & 0.5472 & 0.4780 \\ 
    IRGMcS \cite{Sun2021a} & \underline{98.18} & \underline{0.7791} & \underline{0.7698} & \underline{97.66} & \underline{0.6665} & \underline{0.6544} & \underline{97.86} & \underline{0.8204} & \underline{0.8090} & \underline{93.84} & 0.6456 & 0.6137 & \underline{93.38} & \underline{0.6380} & \underline{0.6067} \\ 
    \textbf{SR-GCAE} & \textbf{98.55} & \textbf{0.8289} & \textbf{0.8214} & \textbf{98.54} & \textbf{0.7760} & \textbf{0.7685} & \textbf{98.70} & \textbf{0.8830} & \textbf{0.8762} & \textbf{98.69} & \textbf{0.9379} & \textbf{0.9306} & \textbf{96.89} & \textbf{0.8634} & \textbf{0.8461} \\ 
    \hline
  \end{tabular}
\end{table*}

\begin{table*}[!ht]
  \setlength\tabcolsep{6pt}
 
  \renewcommand{\arraystretch}{1.4}
  \caption{\centering{Kappa coefficient of different methods on the five multimodal change detection datasets. The accuracy reported here comes from their original papers and is listed ranked from highest to lowest. Our method is highlighted in bold}}
  \label{SOTA_com}
  \centering
  \begin{tabular}{|c c | c c | c c | c c |c c |}
    \hline
    \textbf{Shuguang} &	\textbf{KC} & \textbf{River} & \textbf{KC}
    & \textbf{Farmland} & \textbf{KC} & \textbf{Texas} & \textbf{KC} & \textbf{Hanyang} & \textbf{KC} \\
    \hline        
    \hline
    \textbf{SR-GCAE} &	\textbf{0.8241} &	\textbf{SR-GCAE}	& \textbf{0.7685} &	pt-CDN \cite{Yang2019a}	& 0.8883 &	\textbf{SR-GCAE}	 & \textbf{0.9306} &	\textbf{SR-GCAE} &	\textbf{0.8461}\\
    HGIR-MRF \cite{Sun2022Structured} &	0.779 &	LTFL \cite{Zhan2018Log} &	0.6950 &	DCNet \cite{Gao2019Change} &	0.8833 &	UIR \cite{Luppino2019Unsupervised}	& 0.914	& KPCA-MNet \cite{Wu2021unsupervised}	& 0.8048 \\
    PSGM \cite{Sun2021bPatch}  &	0.7438	& CACD \cite{Wu2021b} &	0.6720	& \textbf{SR-GCAE} &	\textbf{0.8762}    & CAAE \cite{Luppino2022Code}	& 0.885	& DSMSCN \cite{Chen2019Deep}	& 0.7740\\
    CACD \cite{Wu2021b}	&0.7320	&  SCCN \cite{Liu2018}	& 0.6154	&DCNN \cite{Li2019a}	&0.8709&	X-Net \cite{Luppino2022Deep}	&0.767	&SiamCRNN \cite{Chen2019a}&	0.7697\\
    X-Net \cite{Luppino2022Deep}	&0.696	&PCC \cite{Liu2018}	&0.5064	&SARDNN \cite{Gong2016}	&0.8692&	ACE-Net \cite{Luppino2022Deep}	&0.720	&DSFANet \cite{Du2019a}	&0.7556\\
    \cline{3-4} \cline{7-8}
    ACE-Net \cite{Luppino2022Deep}	&0.689&	&	&	RFLICM \cite{Gong2012}&	0.8526	&	& &	OBCD \cite{Wu2021unsupervised}&	0.7174 \\ 
    SCCN \cite{Liu2018}	&0.679	&	&	&	DBN \cite{zhao2014deep}	&0.8495	 	&	& &		USFA \cite{Wu2014}&	0.6530 \\
    cGAN \cite{Niu2019}&	0.662&	&	&	SCCN \cite{Liu2018}&	0.8438& & & ISFA \cite{Wu2014}&	0.6305\\
    \cline{1-2} \cline{5-6} \cline{9-10}
  \end{tabular}
\end{table*}

\begin{figure}[!t]
  \centering
  \includegraphics[width=3.3in]{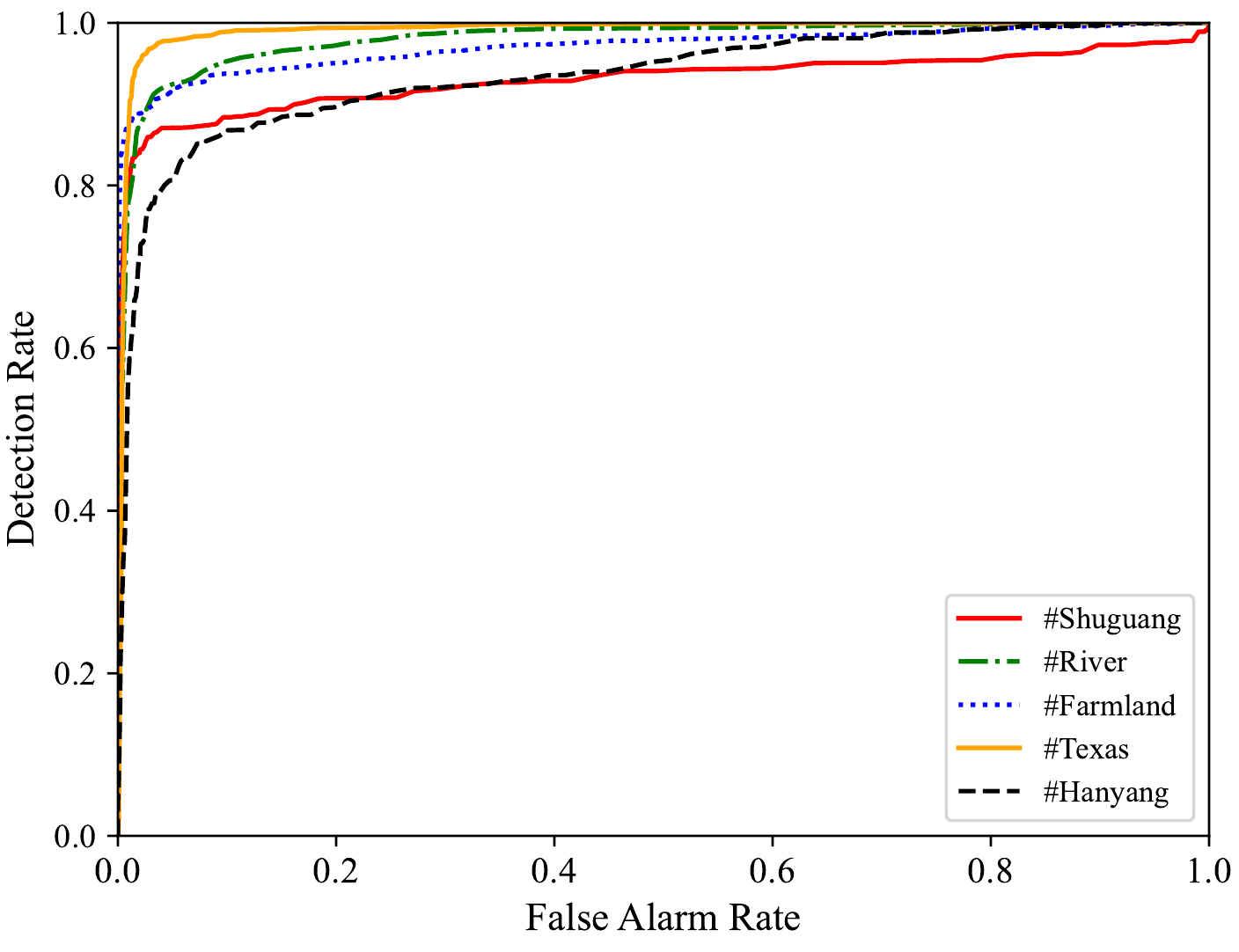}
  \caption{ROC curves of the difference images generated by SR-GCAE on different datasets.}
  \label{roc_curve}
\end{figure}

\subsection{Experiment Settings}
\par The proposed structural relationship graph convolutional autoencoder\footnote{Source code of the proposed method will be available at https://github.com/ChenHongruixuan/SRGCAE} is implemented with the Pytorch library. Adam optimizer with 1$e^{-4}$ learning rate and 1$e^{-6}$ weight decay parameters is applied to optimize the network. In the training process, the maximum number of epochs is set to 20. Two graph convolutional layers with 16 and 32 convolutional kernels are set in the encoder network. In the step of calculating the similarity degree of nonlocal structural relationship, the number of most similar graphs is set to 50. The Otsu algorithm is selected as the threshold segmentation approach to get the binary change map from the generated difference image.

\par To demonstrate the superiority of our method in unsupervised multimodal change detection, we compare it with some state-of-the-art approaches. Firstly, we choose four recently proposed similarity measure-based methods as comparison methods since they are representative and their code is open-sourced:
\begin{enumerate}
    \item M3CD \cite{Touati2020a}. The Markov model for multimodal change detection (M3CD) relies on an observation field built up from a pixel pairwise modeling on heterogeneous image pair. It first estimates the likelihood model parameters using the preliminary iterative estimation technique. Then, the change detection map is computed with a stochastic optimization process.
    \item FPMS \cite{Mignotte2020}. The fractal projection and Markovian segmentation-based method (FPMS) projects the pre-change image to the modality of the post-change image by fractal projection. After the projection, pixel-wise differencing is performed. The difference image is binarized by an MRF segmentation model.
    \item NPSG \cite{Sun2021c}. The non-local patch similarity graph-based method (NPSG) method builds a graph for each image patch based on the self-similarity and then calculates the change level by mapping the graph structure from one image to the other. 
    \item IRGMcS \cite{Sun2021a}. The iterative robust graph and Markovian co-segmentation method (IRG-McS) builds a robust KNN graph to represent the structure of each image and compare the graph to measure the change level. Markovian co-segmentation model is applied to refine the change maps. 
\end{enumerate}

\par Also, we further collect the accuracy obtained by the state-of-the-art (SOTA) methods in each dataset reported in their original paper and list them in Table \ref{SOTA_com}. In this way, the superiority of our method can be fully verified. 

\par In the accuracy assessment step, to evaluate the performance of the change maps generated by the proposed approach and comparison methods, three commonly used evaluation criteria are employed, namely, overall accuracy (OA), F1 score (F1), and Kappa coefficient (KC). In addition, the empirical receiver operating characteristics (ROC) curve is drawn to evaluate the quality of the difference image obtained by our method. The corresponding area under the curve (AUC) is also calculated as the evaluation criterion.

\par All experiments are done on a single PC. The CPU used is Intel Core i7-8750H with a clock rate of 2.2 GHz. The GPU used is a single NVIDIA GeForce GTX 1060.
\begin{figure*}[!ht]
  \centering
  \subfloat[]{
    \includegraphics[height=3.3in]{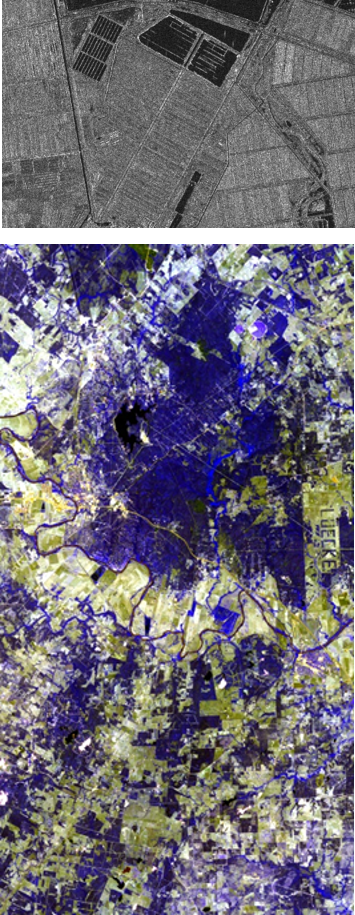}
  \label{fig_first_case}}
  \hfil
  \subfloat[]{
    \includegraphics[height=3.3in]{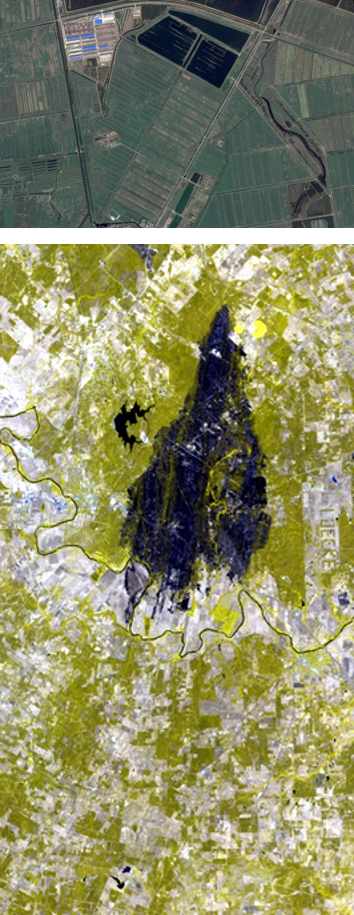}
  \label{fig_second_case}}
  \hfil
  \subfloat[]{
    \includegraphics[height=3.3in]{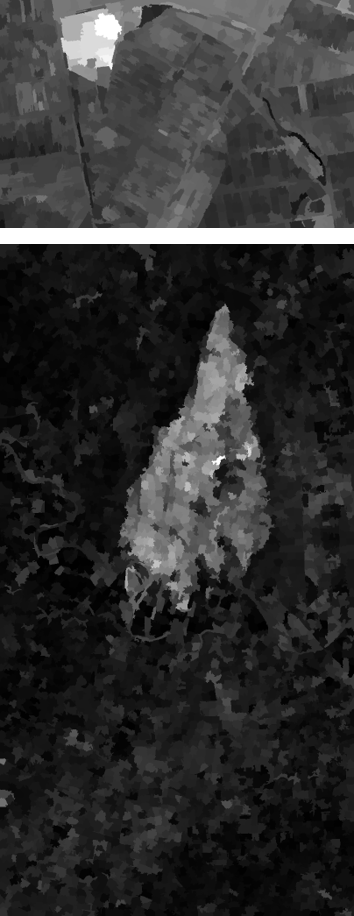}
  \label{fig_third_case}}
  \hfil
 \subfloat[]{
    \includegraphics[height=3.3in]{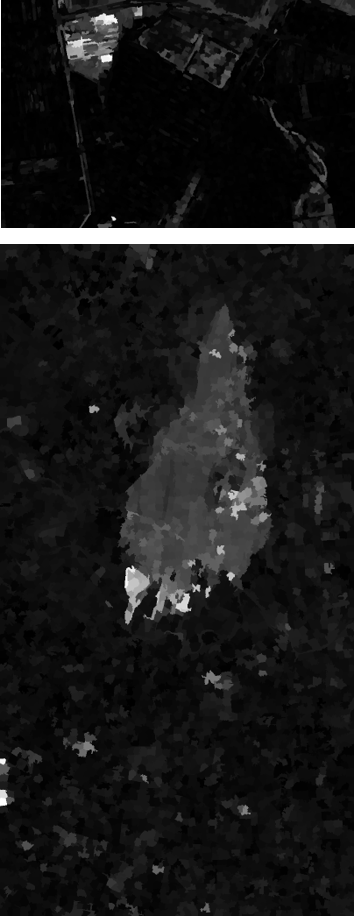}
  \label{fig_second_case}}
  \hfil
   \subfloat[]{
    \includegraphics[height=3.3in]{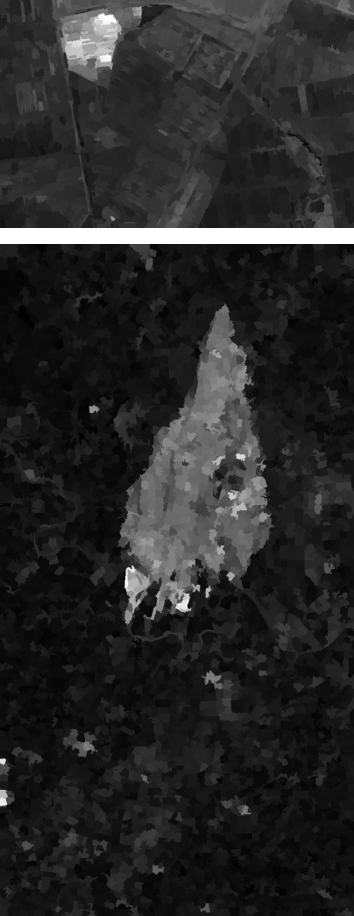}
  \label{fig_second_case}}
  \caption{Difference image obtained by learning different graph information on the two datasets. (a) Pre-change image. (b) Post-change image. (c) Local structural difference image. (d) Nonlocal structural difference image. (e) Fused difference image.}
  \label{di_result}
\end{figure*}

\begin{table*}[!t]
  \setlength\tabcolsep{6pt}
  
  \renewcommand{\arraystretch}{1.4}
  \caption{\centering{The gain in Kappa coefficient of the different step in the proposed framework on five datasets}}
  \label{diff_part_acc}
  \centering
  \begin{tabular}{c c c c c | c c c c c c}
    \hline			
    \multicolumn{5}{c}{\textbf{Method}} & \multicolumn{5}{c}{\textbf{Datasets}}\\
    \hline
    \textbf{\emph{LSR}} & \textbf{\emph{NLSR}} & \textbf{\emph{SR-GCAE}} & \textbf{\emph{Fuse}} & \textbf{\emph{MF}} & \textbf{Shuguang} & \textbf{Farmland}	 & \textbf{River} & \textbf{Texas}	 & \textbf{Hanyang} \\
    \hline\hline
    $\checkmark$& & & &	&0.4053&	0.6811&	0.2548&	0.5453&	0.3173 \\
   $\checkmark$ & & $\checkmark$  & &	&	0.7687&	0.8349&	0.5991&	0.8418&	0.6927\\
    & $\checkmark$& & &	&	0.4325&	0.8270&	0.6178&	0.8111&	0.6550\\
    &$\checkmark$&$\checkmark$& & & 0.5962&	0.8586&	0.7361&	0.8673&	0.8178\\
    $\checkmark$&$\checkmark$&$\checkmark$ &$\checkmark$& & 	0.7923&	0.8729&	0.7466&	0.8950&	0.8213\\
    $\checkmark$&$\checkmark$&$\checkmark$ &$\checkmark$& $\checkmark$&	0.8213&	0.8762&	0.7685&	0.9306&	0.8461\\
    \hline
  \end{tabular}
\end{table*}

\subsection{Change Detection Results}
\par Firstly, the ROC curves of difference images obtained by our method on the five datasets are plotted in Fig. \ref{roc_curve}. We could see that the main areas of distribution for all five curves are on the top left of the figure. The AUC of the five ROC curves are 0.930, 0.980, 0.970, 0.991, and 0.935, respectively. These results suggest that the difference images generated by our method have a very good performance. The changed and unchanged areas would be well separated after the execution of the threshold segmentation algorithm, as shown in Fig. \ref{cd_result}.

\par Fig. \ref{cd_result} shows the binary change maps generated by four comparison methods and SR-GCAE on five multimodal change detection datasets. Due to the mere utilization of low-level information, the four comparison methods cannot obtain accurate change maps on all the datasets. For example, many changed and unchanged areas are not detected by IRGMcS on the Texas dataset. In addition, since these comparison methods are only designed for heterogeneous images, they have difficulty in achieving accurate detection results on the homogeneous Hanyang dataset. In comparison, SR-GCAE obtains accurate change maps with only a few false positive pixels and false negative pixels on all five datasets. Then, Table \ref{acc_ass} lists the OA, F1, and KC of change maps obtained by these different methods. Clearly, SR-GCAE obtains the best values on the three metrics on each dataset. These results demonstrate its effectiveness and practicality in multimodal change detection.

\begin{table}[!t]
  \setlength\tabcolsep{3pt}
  
  \renewcommand{\arraystretch}{1.4}
  \caption{\centering{Accuracy comparison between manual parameter setting and change information adaptive fusion}}
  \label{fusion}
  \centering
  \begin{tabular}{c  c |  c c c c c}
    \hline			
    \multicolumn{2}{c | }{\multirow{2}{*}{\textbf{Method}}} & \multicolumn{5}{c}{\textbf{Datasets}}\\
    \cline{3-7}
     &  &  \textbf{Shuguang} & \textbf{River}	 & \textbf{Farmland} & \textbf{Texas}	 & \textbf{Hanyang} \\
     \hline\hline
    \multirow{5}{*}{\makecell[c]{Manually \\ setting \\ $(\alpha, \beta)$}}	& (1, 0)&	0.7687&	0.6011&	0.8349&	0.8418&	0.6927 \\
    & (0.7, 0.3)&	\underline{0.7909}&	0.6883&	0.8700&	0.8802&	0.7673 \\
    & (0.5, 0.5)&	0.7853&	\underline{0.7420}
    &	\underline{0.8709} &	0.8948&	0.8035 \\
    & (0.3, 0.7)&	0.6391&	0.7362&	0.8683&	\textbf{0.8961} &	\underline{0.8189} \\
    & (0, 1)&	0.5962&	0.7106&	0.8586&	0.8673&	0.8178 \\
    \hline
    \multicolumn{2}{c |}{Adaptive fusion} & \textbf{0.7923} &	\textbf{0.7466} &	\textbf{0.8729}	& \underline{0.8950}	& \textbf{0.8213} \\
    \hline
  \end{tabular}
\end{table}
\par Moreover, to further evaluate the performance of SR-GCAE, we also select SOTA methods for comparison and report the KC obtained by these methods in Table \ref{SOTA_com}. Among these methods, CACD \cite{Wu2021b}, X-Net \cite{Luppino2022Deep}, ACE-Net \cite{Luppino2022Deep}, SCCN \cite{Liu2018}, cGAN \cite{Niu2019}, LTFL \cite{Zhan2018Log}, DCNN \cite{Li2019a}, pt-CDN \cite{Yang2019a}, DCNet \cite{Gao2019Change}, SARDNN \cite{Gong2016}, DBN \cite{zhao2014deep}, CAAE \cite{Luppino2022Code}, KPCA-MNet \cite{Wu2021unsupervised}, DSMSCN \cite{Chen2019Deep}, SiamCRNN \cite{Chen2019a}, and DSFANet \cite{Du2019a} are deep learning-based methods. It can be seen that our method outperforms almost all SOTA methods on all five datasets. Note that even compared to some methods specializing in SAR image change detection, like pt-CDN, and optical image change detection, like KPCA-MNet and DSFANet, SR-GCAE can still show competitive results. The comparison in Table \ref{SOTA_com} further demonstrates the superiority of our method.

\subsection{Discussion}
\par To validate our motivation for introducing graph representation learning and the role of each part in the proposed method, we present the contribution of different steps to the final accuracy in Table \ref{diff_part_acc}. In the Table, \textbf{\emph{LSR}} and \textbf{\emph{NLSR}} refer to only utilizing local and nonlocal structural relationship to detect changes without graph representation learning, respectively. \textbf{\emph{Fuse}} means the adaptive fusion strategy. \textbf{\emph{MF}} indicates the morphological filtering-based post-processing. 

\begin{figure}[!t]
  \centering
  \subfloat[]{
    \includegraphics[width=1.64in]{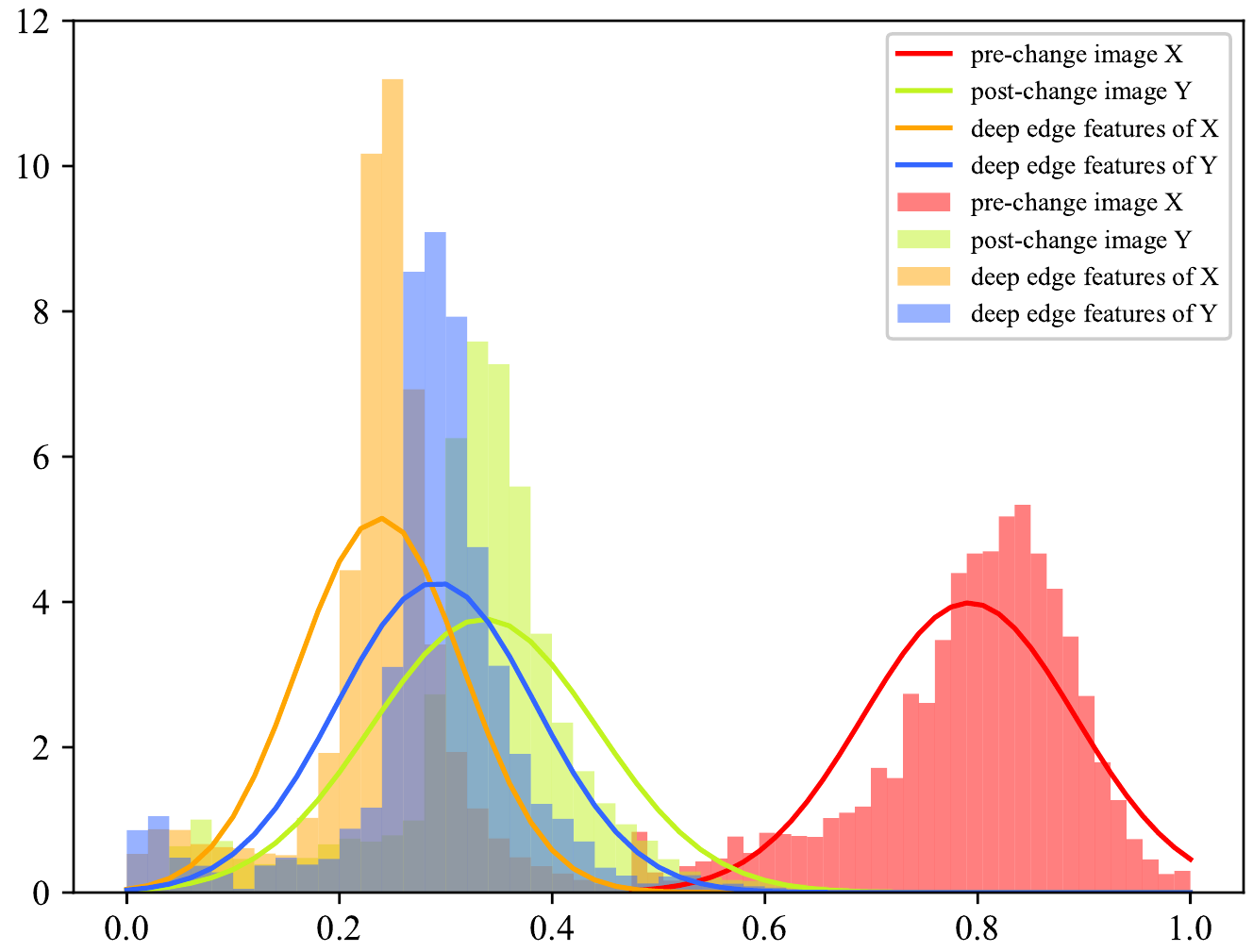}
  \label{fig_first_case}}
  \hfil
  \subfloat[]{
    \includegraphics[width=1.64in]{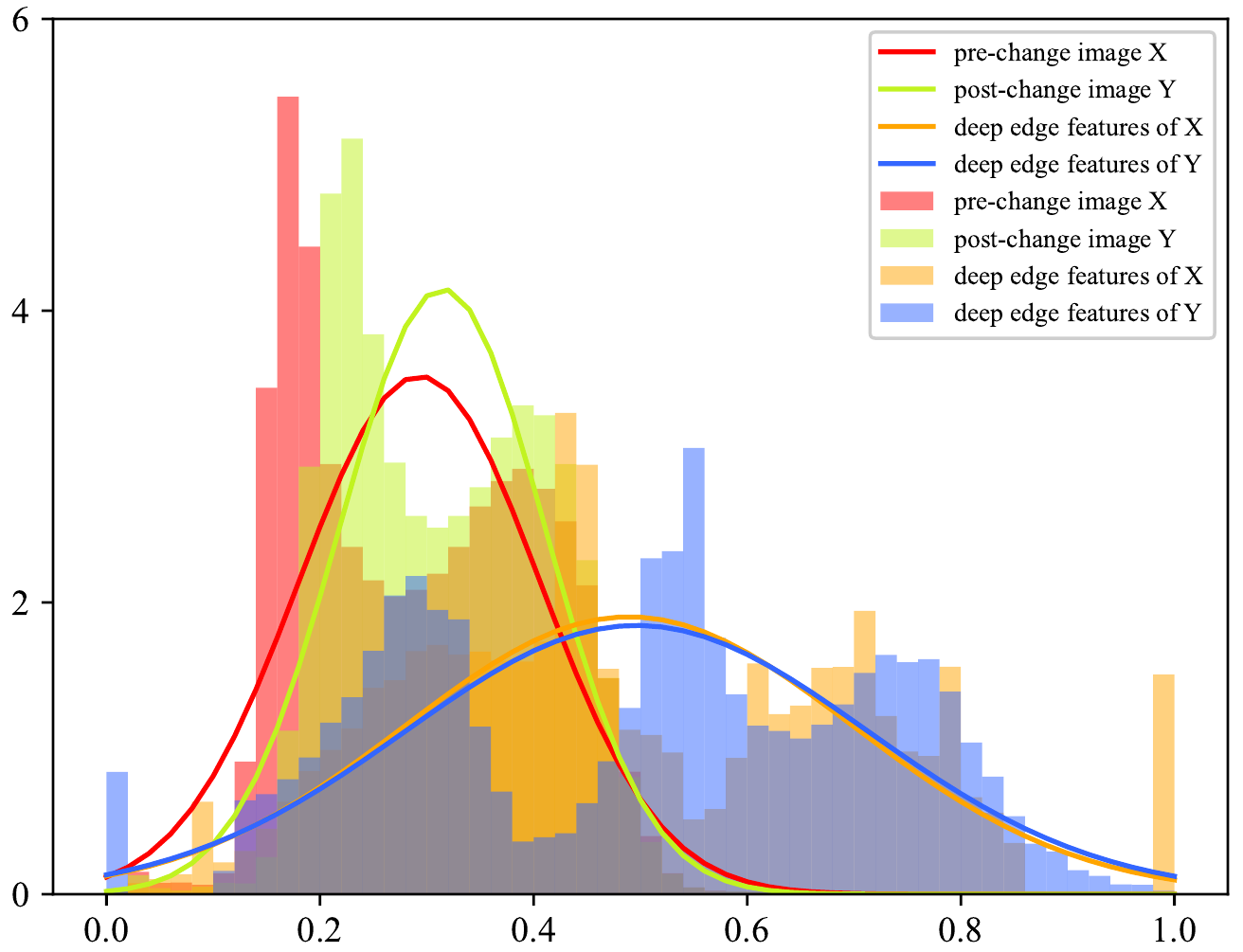}
  \label{fig_second_case}}
  \caption{Distribution comparison of original images and deep edge representation on the (a) Shuguang and (b) Texas datasets.}
  \label{distribution}
\end{figure}

\begin{figure}[ht]
  \centering
  \includegraphics[width=3.3in]{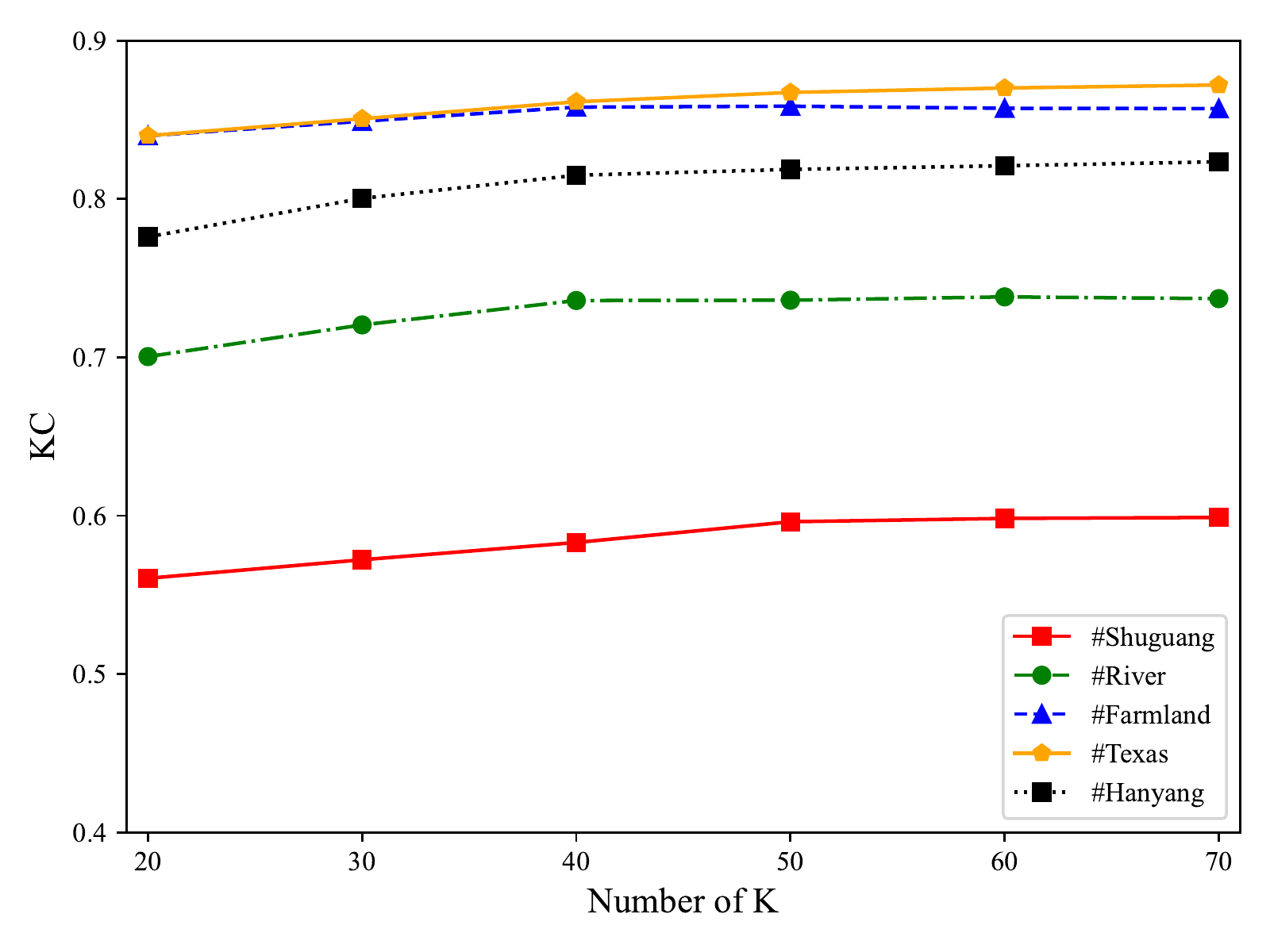}
  \caption{Influences of the number of most similar structural graphs $K$ on the performance of nonlocal structural similarity measurement.}
  \label{fig:Knumber}
\end{figure}

\par Firstly, compared with utilizing low-level information for multimodal change detection, introducing graph representation learning to explore graph information can obviously improve accuracy. For example, in the Shuguang dataset, applying the proposed SR-GCAE to learn vertex and edge information can bring KC increments of 0.3034 and 0.1637 for multimodal change detection. We also visualize the distribution of original data and deep edge representation learned by SR-GCAE on two datasets in Fig. \ref{distribution}. It can be seen that although the distributions of the multimodal data vary considerably in the original spectral domain, they become close after SR-GCAE models the vertex information from structural graphs.
\begin{figure}[!t]
  \centering
  \subfloat[]{
    \includegraphics[height=3.9in]{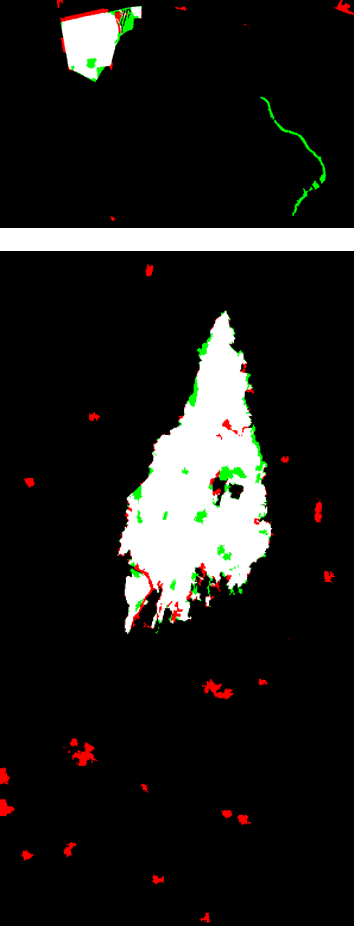}
  \label{fig_first_case}}
  \hfil
  \subfloat[]{
    \includegraphics[height=3.9in]{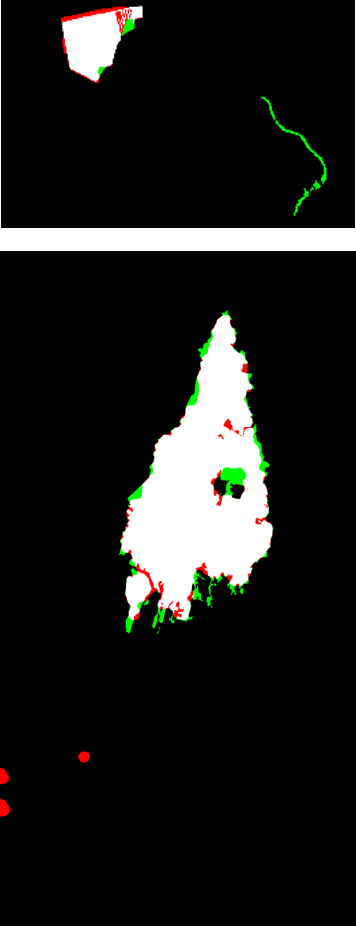}
  \label{fig_second_case}}
  \caption{Change maps (a) before and (b) after morphological filtering. In change maps, White: true positives (TP); Red: false positives (FP); Black: true negatives (TN); Green: false negatives (FN).}
  \label{pps_result}
\end{figure}

\par Then, learning local and nonlocal structure relationships are more suitable on different datasets, respectively. For example, learning local structure relationships can achieve better detection results than learning nonlocal ones on the Shuguang dataset. Besides, the number of most similar graphs $K$ will influence the performance of nonlocal structural similarity measurement. Fig. \ref{fig:Knumber} the effect of $K$ on the accuracy of change detection. It can be seen that a value of around 50 for $K$ is an appropriate choice.

\par After fusing the two difference images, the detection accuracy is further improved. Fig. \ref{di_result} illustrates the local and nonlocal difference images and fused ones. It can be seen that local and nonlocal structural difference images have different distributions of change intensity. After fusion, most of the changed areas are highlighted and most of the unchanged areas are suppressed. Additionally, Table \ref{fusion} compares the performance of our adaptive fusion and manually setting fusion weights $\alpha$ and $\beta$, i.e., $\mathcal{D}\mathcal{I}^{final}=\alpha\mathcal{D}\mathcal{I}^{lcl}+\beta\mathcal{D}\mathcal{I}^{nlcl}$. We could see that the optimal combination of parameters $\alpha$ and $\beta$ varies on different datasets. Although the idea of our fusion method is very simple, it can fuse the two difference images effectively and yields competitive results on all five datasets compared to setting the fusion parameters manually.

\par The final step is morphological filtering. Compared to the Markovian model-based post-processing methods in some approaches, the morphological filtering used in our method is simple and efficient. In five datasets, morphological filtering can further improve the detection performance. Fig. \ref{pps_result} compares the change maps before and after morphological filtering. It can be seen that some noises are eliminated and some changed areas become more intact in refined change maps.

\par Furthermore, the computational overhead of SR-GCAE is evaluated on three datasets with different scales. We report the running time of the four comparison methods and our approach in Table \ref{runtime}. Note that in the C++ code of M3CD and FPMS, the resampling algorithm is performed on the input images to reduce the computational overhead. It can be seen that the running time of our method is an order of magnitude smaller than that of the M3CD and NPSG algorithms. However, FPMS and IRGMcS are more efficient than our method. This running time is acceptable considering the good performance of SR-GCAE and the possibility that we can execute our algorithm on more advanced computers and hardware.

\begin{table}[!t]
  \setlength\tabcolsep{3pt}
  \renewcommand{\arraystretch}{1.4}
  \caption{\centering{Computational overhead of the different methods on three datasets with different scales. The units in the table are in seconds}}
  \label{runtime}
  \centering
  \begin{tabular}{c c | c c c c | c}
    \hline			
    \textbf{Datasets}	&\textbf{Image size}&	\textbf{M3CD}&	\textbf{FPMS}&	\textbf{NSPG}&	\textbf{IRGMcS}	&\textbf{SR-GCAE}\\
    \hline\hline
    Farmland&	306$\times$291	&1178.3	&5.4	&227.3&	3.1	&64.6 \\
    Shuguang&	593$\times$921	&1875.6&	13.2&	1980.8&	29.9&	206.1\\
    Texas&	1534$\times$808&	2316.3&	20.4&	2881.0&	46.4&	580.3 \\
    \hline
  \end{tabular}
\end{table}

\section{Conclusion}\label{sec:5}
\par In this paper, we propose an unsupervised multimodal change detection framework based on two structural relationships of multimodal data and graph convolutional networks. In particular, we construct structural graphs to represent the structural information of multimodal images. Then, we propose a structural relationship graph autoencoder to learn the structural information from constructed graphs. Reconstructing the edge information and vertex information are designed as the optimization objectives of SR-GCAE, respectively so that the learned features can meet the requirement of change detection. Subsequently, we simultaneously explore the local structural relationship and nonlocal structural relationship in multimodal images and utilize the deep graph representations learned by SR-GCAE to detect land-cover changes based on these two relationships. Once the difference images are obtained, a simple and effective fusion strategy based on the variance of change intensity is proposed to fuse the difference images. Finally, a  post-processing method based on morphological filtering is applied to refine the detection result. 

\par The visual and quantities results on four heterogeneous and one homogeneous change detection datasets show that our method outperforms the other competitors, including traditional and deep-learning-based methods. However, the learning stage of SR-GCAE is a bit time-consuming. Therefore, our further work includes but is not limited to reducing the computational overhead of our model to speed up detection.

\ifCLASSOPTIONcaptionsoff
  \newpage
\fi

\bibliographystyle{IEEEtran}
\bibliography{GCNMCD.bib}

\begin{thebibliography}{10}
\providecommand{\url}[1]{#1}
\csname url@samestyle\endcsname
\providecommand{\newblock}{\relax}
\providecommand{\bibinfo}[2]{#2}
\providecommand{\BIBentrySTDinterwordspacing}{\spaceskip=0pt\relax}
\providecommand{\BIBentryALTinterwordstretchfactor}{4}
\providecommand{\BIBentryALTinterwordspacing}{\spaceskip=\fontdimen2\font plus
\BIBentryALTinterwordstretchfactor\fontdimen3\font minus
  \fontdimen4\font\relax}
\providecommand{\BIBforeignlanguage}[2]{{%
\expandafter\ifx\csname l@#1\endcsname\relax
\typeout{** WARNING: IEEEtran.bst: No hyphenation pattern has been}%
\typeout{** loaded for the language `#1'. Using the pattern for}%
\typeout{** the default language instead.}%
\else
\language=\csname l@#1\endcsname
\fi
#2}}
\providecommand{\BIBdecl}{\relax}
\BIBdecl

\bibitem{Singh1989}
A.~Singh, ``{Review Articlel: Digital change detection techniques using
  remotely-sensed data},'' \emph{Int. J. Remote Sens.}, vol.~10, no.~6, pp.
  989--1003, 1989.

\bibitem{Desclee2006}
B.~Descl{\'{e}}e, P.~Bogaert, and P.~Defourny, ``{Forest change detection by
  statistical object-based method},'' \emph{Remote Sens. Environ.}, vol. 102,
  no. 1-2, pp. 1--11, 2006.

\bibitem{Brunner2010}
D.~Brunner, G.~Lemoine, and L.~Bruzzone, ``{Earthquake damage assessment of
  buildings using VHR optical and SAR imagery},'' \emph{IEEE Trans. Geosci.
  Remote Sens.}, vol.~48, no.~5, pp. 2403--2420, 2010.

\bibitem{Wu2017b}
C.~Wu, B.~Du, X.~Cui, and L.~Zhang, ``{A post-classification change detection
  method based on iterative slow feature analysis and Bayesian soft fusion},''
  \emph{Remote Sens. Environ.}, vol. 199, pp. 241--255, 2017.

\bibitem{Luo2018}
H.~Luo, C.~Liu, C.~Wu, and X.~Guo, ``{Urban change detection based on
  Dempster-Shafer theory for multitemporal very high-resolution imagery},''
  \emph{Remote Sensing}, vol.~10, no.~7, pp. 20--22, 2018.

\bibitem{Wu2021a}
C.~Wu, Y.~Guo, H.~Guo, J.~Yuan, L.~Ru, H.~Chen, B.~Du, and L.~Zhang, ``{An
  Investigation of Traffic Density Changes inside Wuhan during the COVID-19
  Epidemic with GF-2 Time-Series Images},'' \emph{Int. J. Appl. Earth Obs.
  Geoinf.}, vol. 103, p. 102503, 2021.

\bibitem{Chen2022Dual}
H.~Chen, E.~Nemni, S.~Vallecorsa, X.~Li, C.~Wu, and L.~Bromley, ``Dual-tasks
  siamese transformer framework for building damage assessment,'' in
  \emph{International Geoscience and Remote Sensing Symposium (IGARSS)}, 2022,
  pp. 1--4.

\bibitem{Sharma2007}
L.~Bruzzone and {Diego Fern{\`{a}}ndez Prieto}, ``{Automatic Analysis of the
  Difference Image for Unsupervised Change Detection},'' \emph{IEEE Trans.
  Geosci. Remote Sens.}, vol.~38, no.~3, pp. 1171--1182, 2000.

\bibitem{Wu2014}
C.~Wu, B.~Du, and L.~Zhang, ``{Slow feature analysis for change detection in
  multispectral imagery},'' \emph{IEEE Trans. Geosci. Remote Sens.}, vol.~52,
  no.~5, pp. 2858--2874, 2014.

\bibitem{Du2019a}
B.~Du, L.~Ru, C.~Wu, and L.~Zhang, ``{Unsupervised Deep Slow Feature Analysis
  for Change Detection in Multi-Temporal Remote Sensing Images},'' \emph{IEEE
  Trans. Geosci. Remote Sens.}, vol.~57, no.~12, pp. 9976--9992, 2019.

\bibitem{Mou2019}
L.~Mou, L.~Bruzzone, and X.~X. Zhu, ``{Learning spectral-spatialoral features
  via a recurrent convolutional neural network for change detection in
  multispectral imagery},'' \emph{IEEE Trans. Geosci. Remote Sens.}, vol.~57,
  no.~2, pp. 924--935, 2019.

\bibitem{Liu2019}
S.~Liu, D.~Marinelli, L.~Bruzzone, and F.~Bovolo, ``{A review of change
  detection in multitemporal hyperspectral images: Current techniques,
  applications, and challenges},'' \emph{IEEE Geosci. Remote Sens. Mag.},
  vol.~7, no.~2, pp. 140--158, 2019.

\bibitem{Liu2015a}
S.~Liu, L.~Bruzzone, and F.~Bovolo, ``{Hierarchical Unsupervised Change
  Detection in Multitemporal Hyperspectral Images},'' \emph{IEEE Trans. Geosci.
  Remote Sens.}, vol.~53, no.~1, pp. 244--260, 2015.

\bibitem{Liu2015b}
S.~Liu, L.~Bruzzone, F.~Bovolo, M.~Zanetti, and P.~Du, ``{Sequential Spectral
  Change Vector Analysis for Iteratively Discovering and Detecting Multiple
  Changes in Hyperspectral Images},'' \emph{IEEE Trans. Geosci. Remote Sens.},
  vol.~53, no.~8, pp. 4363--4378, 2015.

\bibitem{Wang2019}
Q.~Wang, Z.~Yuan, Q.~Du, and X.~Li, ``{GETNET: A General End-To-End 2-D CNN
  Framework for Hyperspectral Image Change Detection},'' \emph{IEEE Trans.
  Geosci. Remote Sens.}, vol.~57, no.~1, pp. 3--13, 2019.

\bibitem{Hu2021Hyperspectral}
M.~Hu, C.~Wu, L.~Zhang, and B.~Du, ``Hyperspectral anomaly change detection
  based on autoencoder,'' \emph{IEEE J. Sel. Top. Appl. Earth Obs. Remote
  Sens.}, vol.~14, pp. 3750--3762, 2021.

\bibitem{Chen2019Deep}
H.~Chen, C.~Wu, B.~Du, and L.~Zhang, ``{Deep Siamese Multi-scale Convolutional
  Network for Change Detection in Multi-Temporal VHR Images},'' in \emph{2019
  10th International Workshop on the Analysis of Multitemporal Remote Sensing
  Images, MultiTemp 2019}, 2019, pp. 1--4.

\bibitem{Saha2019}
S.~Saha, F.~Bovolo, and L.~Bruzzone, ``{Unsupervised deep change vector
  analysis for multiple-change detection in VHR Images},'' \emph{IEEE Trans.
  Geosci. Remote Sens.}, vol.~57, no.~6, pp. 3677--3693, 2019.

\bibitem{Peng2020}
D.~Peng, L.~Bruzzone, Y.~Zhang, H.~Guan, H.~Ding, and X.~Huang, ``{SemiCDNet :
  A Semisupervised Convolutional Neural Network for Change Detection in High
  Resolution Remote-Sensing Images},'' \emph{IEEE Trans. Geosci. Remote Sens.},
  pp. 1--16, 2020.

\bibitem{Wu2021unsupervised}
C.~Wu, H.~Chen, B.~Du, L.~Zhang, B.~Do, and L.~Zhang, ``{Unsupervised Change
  Detection in Multitemporal VHR Images Based on Deep Kernel PCA Convolutional
  Mapping Network},'' \emph{IEEE Trans. Cybern}, pp. 1--15, 2021.

\bibitem{guo2021deep}
H.~Guo, Q.~Shi, A.~Marinoni, B.~Du, and L.~Zhang, ``Deep building footprint
  update network: A semi-supervised method for updating existing building
  footprint from bi-temporal remote sensing images,'' \emph{Remote Sens.
  Environ.}, vol. 264, p. 112589, 2021.

\bibitem{Tang2022}
X.~Tang, H.~Zhang, L.~Mou, F.~Liu, X.~Zhang, X.~X. Zhu, and L.~Jiao, ``{An
  Unsupervised Remote Sensing Change Detection Method Based on Multiscale Graph
  Convolutional Network and Metric Learning},'' \emph{IEEE Trans. Geosci.
  Remote Sens.}, vol.~60, 2022.

\bibitem{Gong2012}
M.~Gong, Z.~Zhou, and J.~Ma, ``{Change detection in synthetic aperture radar
  images based on image fusion and fuzzy clustering},'' \emph{IEEE Trans. Image
  Process.}, vol.~21, no.~4, pp. 2141--2151, 2012.

\bibitem{Gong2016}
M.~Gong, J.~Zhao, J.~Liu, Q.~Miao, and L.~Jiao, ``{Change Detection in
  Synthetic Aperture Radar Images Based on Deep Neural Networks},'' \emph{IEEE
  Trans. Neural Netw. Learn. Syst.}, vol.~27, no.~1, pp. 125--138, 2016.

\bibitem{Gao2016}
F.~Gao, J.~Dong, B.~Li, and Q.~Xu, ``{Automatic Change Detection in Synthetic
  Aperture Radar Images Based on PCANet},'' \emph{IEEE Geosci. Remote Sens.
  Lett.}, vol.~13, no.~12, pp. 1792--1796, 2016.

\bibitem{Yang2019a}
M.~Yang, L.~Jiao, F.~Liu, B.~Hou, and S.~Yang, ``{Transferred deep
  learning-based change detection in remote sensing images},'' \emph{IEEE
  Transactions on Geoscience and Remote Sensing}, vol.~57, no.~9, pp.
  6960--6973, 2019.

\bibitem{Sun2020}
Y.~Sun, L.~Lei, D.~Guan, X.~Li, and G.~Kuang, ``{SAR Image Change Detection
  Based on Nonlocal Low-Rank Model and Two-Level Clustering},'' \emph{IEEE J.
  Sel. Top. Appl. Earth Obs. Remote Sens.}, vol.~13, pp. 293--306, 2020.

\bibitem{Chen2019a}
H.~Chen, C.~Wu, B.~Du, L.~Zhang, and L.~Wang, ``{Change Detection in
  Multisource VHR Images via Deep Siamese Convolutional Multiple-Layers
  Recurrent Neural Network},'' \emph{IEEE Trans. Geosci. Remote Sens.},
  vol.~58, no.~4, pp. 2848--2864, 2020.

\bibitem{Wu2021multiscale}
J.~Wu, B.~Li, Y.~Qin, W.~Ni, H.~Zhang, R.~Fu, and Y.~Sun, ``{A multiscale graph
  convolutional network for change detection in homogeneous and heterogeneous
  remote sensing images},'' \emph{Int. J. Appl. Earth Obs. Geoinf.}, vol. 105,
  p. 102615, 2021.

\bibitem{Wan2019Post}
L.~Wan, Y.~Xiang, and H.~You, ``{A Post-Classification Comparison Method for
  SAR and Optical Images Change Detection},'' \emph{IEEE Geosci. Remote Sens.
  Lett.}, vol.~16, no.~7, pp. 1026--1030, 2019.

\bibitem{Wan2019An}
------, ``{An object-based hierarchical compound classification method for
  change detection in heterogeneous optical and SAR images},'' \emph{IEEE
  Trans. Geosci. Remote Sens.}, vol.~57, no.~12, pp. 9941--9959, 2019.

\bibitem{Niu2019}
X.~Niu, M.~Gong, T.~Zhan, and Y.~Yang, ``{A Conditional Adversarial Network for
  Change Detection in Heterogeneous Images},'' \emph{IEEE Geosci. Remote Sens.
  Lett.}, vol.~16, no.~1, pp. 45--49, 2019.

\bibitem{Jiang2020Change}
X.~Jiang, G.~Li, Y.~Liu, X.~P. Zhang, and Y.~He, ``{Change Detection in
  Heterogeneous Optical and SAR Remote Sensing Images Via Deep Homogeneous
  Feature Fusion},'' \emph{Sel.}, vol.~13, pp. 1551--1566, 2020.

\bibitem{Luppino2022Code}
L.~T. Luppino, M.~A. Hansen, M.~Kampffmeyer, F.~M. Bianchi, G.~Moser,
  R.~Jenssen, and S.~N. Anfinsen, ``Code-aligned autoencoders for unsupervised
  change detection in multimodal remote sensing images,'' \emph{IEEE Trans.
  Neural Netw. Learn. Syst.}, pp. 1--13, 2022.

\bibitem{Luppino2022Deep}
L.~T. Luppino, M.~Kampffmeyer, F.~M. Bianchi, G.~Moser, S.~B. Serpico,
  R.~Jenssen, and S.~N. Anfinsen, ``{Deep Image Translation with an
  Affinity-Based Change Prior for Unsupervised Multimodal Change Detection},''
  \emph{IEEE Trans. Geosci. Remote Sens.}, vol.~60, 2022.

\bibitem{Zhang2016Change}
P.~Zhang, M.~Gong, L.~Su, J.~Liu, and Z.~Li, ``{Change detection based on deep
  feature representation and mapping transformation for
  multi-spatial-resolution remote sensing images},'' \emph{ISPRS J. Photogramm.
  Remote Sens.}, vol. 116, pp. 24--41, 2016.

\bibitem{Liu2018}
J.~Liu, M.~Gong, K.~Qin, and P.~Zhang, ``{A Deep Convolutional Coupling Network
  for Change Detection Based on Heterogeneous Optical and Radar Images},''
  \emph{IEEE Trans. Neural Netw. Learn. Syst.}, vol.~29, no.~3, pp. 545--559,
  2018.

\bibitem{Zhao2017a}
W.~Zhao, Z.~Wang, M.~Gong, and J.~Liu, ``{Discriminative Feature Learning for
  Unsupervised Change Detection in Heterogeneous Images Based on a Coupled
  Neural Network},'' \emph{IEEE Trans. Geosci. Remote Sens.}, vol.~55, no.~12,
  pp. 7066--7080, 2017.

\bibitem{Zhan2018b}
T.~Zhan, M.~Gong, J.~Liu, and P.~Zhang, ``{Iterative feature mapping network
  for detecting multiple changes in multi- source remote sensing images},''
  \emph{ISPRS J. Photogramm. Remote Sens.}, vol. 146, pp. 38--51, 2018.

\bibitem{Zhan2018Log}
T.~Zhan, M.~Gong, X.~Jiang, and S.~Li, ``{Log-based transformation feature
  learning for change detection in heterogeneous images},'' \emph{IEEE Geosci.
  Remote Sens. Lett.}, vol.~15, no.~9, pp. 1352--1356, 2018.

\bibitem{Wu2021b}
Y.~Wu, J.~Li, Y.~Yuan, A.~K. Qin, Q.~G. Miao, and M.~G. Gong, ``{Commonality
  Autoencoder: Learning Common Features for Change Detection From Heterogeneous
  Images},'' \emph{IEEE Trans. Neural Netw. Learn. Syst.}, pp. 1--14, 2021.

\bibitem{Wan2018Multi}
L.~Wan, T.~Zhang, and H.~J. You, ``{Multi-sensor remote sensing image change
  detection based on sorted histograms},'' \emph{Int. J. Remote Sens.},
  vol.~39, no.~11, pp. 3753--3775, 2018.

\bibitem{Liu2018a}
Z.~Liu, G.~Li, G.~Mercier, Y.~He, and Q.~Pan, ``{Change Detection in
  Heterogenous Remote Sensing Images via Homogeneous Pixel Transformation},''
  \emph{IEEE Trans. Image Process.}, vol.~27, no.~4, pp. 1822--1834, 2018.

\bibitem{Luppino2019Unsupervised}
L.~T. Luppino, F.~M. Bianchi, G.~Moser, and S.~N. Anfinse, ``{Unsupervised
  Image Regression for Heterogeneous Change Detection},'' \emph{IEEE Trans.
  Geosci. Remote Sens.}, vol.~57, no.~12, pp. 9960--9975, 2019.

\bibitem{Mignotte2020}
M.~Mignotte, ``{A Fractal Projection and Markovian Segmentation-Based Approach
  for Multimodal Change Detection},'' \emph{IEEE Trans. Geosci. Remote Sens.},
  vol.~58, no.~11, pp. 8046--8058, 2020.

\bibitem{Sun2021bPatch}
Y.~Sun, L.~Lei, X.~Li, X.~Tan, and G.~Kuang, ``{Patch Similarity Graph
  Matrix-Based Unsupervised Remote Sensing Change Detection with Homogeneous
  and Heterogeneous Sensors},'' \emph{IEEE Trans. Geosci. Remote Sens.},
  vol.~59, no.~6, pp. 4841--4861, 2021.

\bibitem{Sun2021c}
Y.~Sun, L.~Lei, X.~Li, H.~Sun, and G.~Kuang, ``{Nonlocal patch similarity based
  heterogeneous remote sensing change detection},'' \emph{Pattern Recognition},
  vol. 109, pp. 1--16, 2021.

\bibitem{Sun2021a}
Y.~Sun, L.~Lei, D.~Guan, and G.~Kuang, ``{Iterative Robust Graph for
  Unsupervised Change Detection of Heterogeneous Remote Sensing Images},''
  \emph{IEEE Trans. Image Process.}, vol.~30, pp. 6277--6291, 2021.

\bibitem{Buades2005non}
A.~Buades, B.~Coll, and J.-M. Morel, ``A non-local algorithm for image
  denoising,'' in \emph{2005 IEEE Computer Society Conference on Computer
  Vision and Pattern Recognition (CVPR'05)}, vol.~2, 2005, pp. 60--65 vol. 2.

\bibitem{dabov2007image}
K.~Dabov, A.~Foi, V.~Katkovnik, and K.~Egiazarian, ``Image denoising by sparse
  3-d transform-domain collaborative filtering,'' \emph{IEEE Trans. Image
  Process.}, vol.~16, no.~8, pp. 2080--2095, 2007.

\bibitem{buades2010image}
A.~Buades, B.~Coll, and J.-M. Morel, ``Image denoising methods. a new nonlocal
  principle,'' \emph{SIAM review}, vol.~52, no.~1, pp. 113--147, 2010.

\bibitem{kipf2016semi}
T.~N. Kipf and M.~Welling, ``Semi-supervised classification with graph
  convolutional networks,'' \emph{arXiv preprint arXiv:1609.02907}, 2016.

\bibitem{defferrard2016convolutional}
M.~Defferrard, X.~Bresson, and P.~Vandergheynst, ``Convolutional neural
  networks on graphs with fast localized spectral filtering,'' \emph{Advances
  in neural information processing systems}, vol.~29, 2016.

\bibitem{velivckovic2017graph}
P.~Veli{\v{c}}kovi{\'c}, G.~Cucurull, A.~Casanova, A.~Romero, P.~Lio, and
  Y.~Bengio, ``Graph attention networks,'' \emph{arXiv preprint
  arXiv:1710.10903}, 2017.

\bibitem{wu2019simplifying}
F.~Wu, A.~Souza, T.~Zhang, C.~Fifty, T.~Yu, and K.~Weinberger, ``Simplifying
  graph convolutional networks,'' in \emph{International conference on machine
  learning}.\hskip 1em plus 0.5em minus 0.4em\relax PMLR, 2019, pp. 6861--6871.

\bibitem{park2019symmetric}
J.~Park, M.~Lee, H.~J. Chang, K.~Lee, and J.~Y. Choi, ``Symmetric graph
  convolutional autoencoder for unsupervised graph representation learning,''
  in \emph{Proceedings of the IEEE/CVF International Conference on Computer
  Vision}, 2019, pp. 6519--6528.

\bibitem{hammond2011wavelets}
D.~K. Hammond, P.~Vandergheynst, and R.~Gribonval, ``Wavelets on graphs via
  spectral graph theory,'' \emph{Applied and Computational Harmonic Analysis},
  vol.~30, no.~2, pp. 129--150, 2011.

\bibitem{Zitova2003}
B.~Zitov{\'{a}} and J.~Flusser, ``{Image registration methods: A survey},''
  \emph{Image and Vision Computing}, vol.~21, no.~11, pp. 977--1000, 2003.

\bibitem{Zhang2014}
L.~Zhang, C.~Wu, and B.~Du, ``{Automatic radiometric normalization for
  multitemporal remote sensing imagery with iterative slow feature analysis},''
  \emph{IEEE Trans. Geosci. Remote Sens.}, vol.~52, no.~10, pp. 6141--6155,
  2014.

\bibitem{Li2019a}
Y.~Li, C.~Peng, Y.~Chen, L.~Jiao, L.~Zhou, and R.~Shang, ``{A Deep Learning
  Method for Change Detection in Synthetic Aperture Radar Images},'' \emph{IEEE
  Trans. Geosci. Remote Sens.}, vol.~57, no.~8, pp. 5751--5763, 2019.

\bibitem{baatz2000multi}
M.~Baatz, ``Multi resolution segmentation: an optimum approach for high quality
  multi scale image segmentation,'' in \emph{Beutrage zum AGIT-Symposium.
  Salzburg, Heidelberg, 2000}, 2000, pp. 12--23.

\bibitem{Achanta2012SLIC}
R.~Achanta, A.~Shaji, K.~Smith, A.~Lucchi, P.~Fua, and S.~Süsstrunk, ``Slic
  superpixels compared to state-of-the-art superpixel methods,'' \emph{IEEE
  Transactions on Pattern Analysis and Machine Intelligence}, vol.~34, no.~11,
  pp. 2274--2282, 2012.

\bibitem{Bovolo2007a}
F.~Bovolo and L.~Bruzzone, ``{A Theoretical Framework for Unsupervised Change
  Detection Based on Change Vector Analysis in the Polar Domain},'' \emph{IEEE
  Trans. Geosci. Remote Sens.}, vol.~45, no.~1, pp. 218--236, 2007.

\bibitem{Liu2019c}
G.~Liu, L.~Li, L.~Jiao, Y.~Dong, and X.~Li, ``{Stacked Fisher autoencoder for
  SAR change detection},'' \emph{Pattern Recognition}, vol.~96, p. 106971,
  2019.

\bibitem{Mou2018}
L.~Mou, P.~Ghamisi, and X.~X. Zhu, ``{Unsupervised spectral-spatial feature
  learning via deep residual conv-deconv network for hyperspectral image
  classification},'' \emph{IEEE Trans. Geosci. Remote Sens.}, vol.~56, no.~1,
  pp. 391--406, 2018.

\bibitem{otsu1979}
N.~Otsu, ``A threshold selection method from gray-level histograms,''
  \emph{IEEE transactions on systems, man, and cybernetics}, vol.~9, no.~1, pp.
  62--66, 1979.

\bibitem{moon1996}
T.~K. Moon, ``The expectation-maximization algorithm,'' \emph{IEEE Signal
  processing magazine}, vol.~13, no.~6, pp. 47--60, 1996.

\bibitem{Touati2020a}
R.~Touati, M.~Mignotte, and M.~Dahmane, ``{Multimodal Change Detection in
  Remote Sensing Images Using an Unsupervised Pixel Pairwise-Based Markov
  Random Field Model},'' \emph{IEEE Trans. Image Process.}, vol.~29, pp.
  757--767, 2020.

\bibitem{Sun2022Structured}
Y.~Sun, L.~Lei, X.~Tan, D.~Guan, J.~Wu, and G.~Kuang, ``{Structured graph based
  image regression for unsupervised multimodal change detection},'' \emph{ISPRS
  J. Photogramm. Remote Sens.}, vol. 185, pp. 16--31, 2022.

\bibitem{Gao2019Change}
Y.~Gao, F.~Gao, J.~Dong, and S.~Wang, ``{Change Detection From Synthetic
  Aperture Radar Images Based on Channel Weighting-Based Deep Cascade
  Network},'' \emph{IEEE J. Sel. Top. Appl. Earth Obs. Remote Sens.}, vol.~12,
  no.~11, pp. 4517--4529, 2019.

\bibitem{zhao2014deep}
J.~Zhao, M.~Gong, J.~Liu, and L.~Jiao, ``Deep learning to classify difference
  image for image change detection,'' in \emph{2014 International Joint
  Conference on Neural Networks (IJCNN)}.\hskip 1em plus 0.5em minus
  0.4em\relax IEEE, 2014, pp. 411--417.

\end{thebibliography}

\end{document}